\documentclass[runningheads]{llncs}
\usepackage{graphicx}
\usepackage{color}
\usepackage{times}
\usepackage{authblk}
\usepackage{amsmath}
\usepackage{bbding}
\usepackage{amsfonts}
\usepackage{mathtools}
\usepackage{float}
\usepackage{subfigure}
\usepackage{wrapfig}
\usepackage{ulem}
\usepackage{changepage}
\usepackage{svg}
\usepackage{hyperref}
\graphicspath{{figures/}}
\usepackage{authblk}
\usepackage{color}
\begin{document}
\title{Generating 3D Brain Tumor Regions in MRI using Vector-Quantization Generative Adversarial Networks}
\titlerunning{Generating 3D Brain Tumor ROIs using VQGAN}
\author{Meng Zhou\inst{1,2} \and
Matthias W Wagner\inst{3,9} \and
Uri Tabori \inst{4} \and
Cynthia Hawkins \inst{5} \and
Birgit B Ertl-Wagner \inst{2,3,6,7} \and
Farzad Khalvati \inst{1,2,3,6,7,8}\thanks{Corresponding Author: farzad.khalvati@utoronto.ca}}
%
%\author{Meng Zhou}
\authorrunning{M. Zhou et al.}
\institute{Department of Computer Science, University of Toronto \and
Neurosciences \& Mental Health Research Program, The Hospital for Sick Children \and
Department of Diagnostic and Interventional Radiology, The Hospital for Sick Children \and
Division of Neuroradiology, Neurooncology, The Hospital for Sick Children \and
Paediatric Laboratory Medicine, Division of Pathology, The Hospital for Sick Children \and
Institute of Medical Science, University of Toronto \and
Department of Medical Imaging, University of Toronto \and
Department of Mechanical and Industrial Engineering, University of Toronto \and
Department of Neuroradiology, University Hospital Augsburg, Germany \\}
%\institute{Anonymous Organization \\ \email{**@******.***}}
\maketitle              % typeset the header of the contribution
\begin{abstract}
% Deep learning has revealed an unprecedented trend in medical image analysis in recent years, yet training such models often requires a vast amount of data. It is often infeasible to collect large medical imaging datasets under real clinical settings. While GANs have been recently used to augment training data by generating realistic and diverse images, most of these methods require a large amount of data to train can generate realistic and diverse 3D brain tumor ROIs with high-resolution based on vector-quantization GAN and transformer. Our generated samples can be directly used as augmented data in brain tumor ROI-based classification. the classifier outperforms the baseline model (the naive approach without using any augmentation techinques) by over 10\% (73.2\%  vs. 62.1\%) in area under ROC curve. Our results reveal that the potential for GANs to generate tumor ROIs for imbalanced data classification from a small set of training data.  Research on deep learning-based brain tumor classification using MRI has shown that it is easier to classify the ROIs (tumor regions) compared to the entire image volume.

Medical image analysis has significantly benefited from advancements in deep learning, particularly in the application of Generative Adversarial Networks (GANs) for generating realistic and diverse images that can augment training datasets. However, the effectiveness of such approaches is often limited by the amount of available data in clinical settings. Additionally, the common GAN-based approach is to generate entire image volumes, rather than solely the region of interest (ROI) such as the tumor region. Research on deep learning-based brain tumor classification using MRI has shown that it is easier to classify the tumor ROIs compared to the entire image volumes. In this work, we present a novel framework that uses vector-quantization GAN and a transformer incorporating masked token modeling to generate high-resolution and diverse 3D brain tumor ROIs that can be directly used as augmented data for the classification of brain tumor ROI. We apply our method to two imbalanced datasets where we augment the minority class: (1) the Multimodal Brain Tumor Segmentation Challenge (BraTS) 2019 dataset to generate new low-grade glioma (LGG) ROIs to balance with high-grade glioma (HGG) class; (2) the internal pediatric LGG (pLGG) dataset tumor ROIs with BRAF V600E Mutation genetic marker to balance with BRAF Fusion genetic marker class. We show that the proposed method outperforms various baseline models in both qualitative and quantitative measurements. The generated data was used to tackle the problem of imbalanced data in the brain tumor types classification task. Using the augmented data, our approach demonstrates superior performance, surpassing baseline models by up to 6.4\% in the area under the ROC curve (AUC), 3.4\% in F1-score, and 5.4\% in Accuracy in the BraTS 2019 dataset; 4.3\% in AUC, 7.3\% in F1-score, and 9.2\% in Accuracy on our internal pLGG dataset. The results indicate the generated tumor ROIs can effectively address the imbalanced data problem. Our proposed method has the potential to facilitate an accurate diagnosis of rare brain tumors using MRI scans.
%indicating the potential of GAN-based methods to generate tumor ROIs for imbalanced data classification from a small set of training data. Our method has the potential to facilitate more accurate diagnosis of brain tumors in MRI scans.

\end{abstract}
\section{Introduction}
% {\color{red} outline}

% \begin{enumerate}
%     \item Brain tumor is one of the severe diseases among people, talk about HGG and LGG among adults
%     \item LGG is relatively rare than HGG, mention limited number of patients in LGG
%     \item deep learning relies heavily on large amount of data, imbalanced data cause problem
%     \item talk about 3D data, the depth direction-relationship, so less data than 2D
%     \item tumor classification using 3D ROI, generating 3D data is hard, as well as generating seg mask
%     \item contribution: first attempt to use one of the SOTA algo: vqgan to generate 3D brain tumor ROI to use in imbalanced classification task
    
% \end{enumerate}

%Brain tumor is one of most severe diseases among adults ismael2018brain
As one of the most frequent primary brain tumor types within the central nervous system among adults \cite{menze2014multimodal}, gliomas commonly arise from the glial cells and then spread out to the surrounding tissues \cite{holland2001progenitor}. Among all variations of gliomas, the high-grade glioma (HGG), such as glioblastomas and anaplasticgliomas \cite{villa20182016,wang2013understanding} accounts for the majority of cases ($60\% - 75\%$), and patients who have been diagnosed with HGG have a relatively low median survival time of two years or less and require immediate treatment \cite{menze2014multimodal,wang2013understanding}. Another variation (minority cases) is the low-grade glioma (LGG), such as astrocytomas and oligodendrogliomas \cite{youssef2020lower}. This type of glioma is rare, hard to cure, and frequently transforms to the HGG within its lifetime \cite{youssef2020lower}. For both variations, intensive neuroimaging scans are acquired before diagnosis and after the treatment to evaluate the progression of the disease \cite{menze2014multimodal}. In a real-world clinical routine, the commonly used technique for neuroimaging is the multi-parametric Magnetic Resonance Imaging (MRI) equipped with traditional morphologic sequences such as standard T2 or T1-weighted, and other functional forms of imaging such as the Fluid Attenuated Inversion Recovery (FLAIR), and diffusion-weighted MRI \cite{menze2014multimodal}. Each of these modalities provides different biological information about the tumor and can be used by radiologists to diagnose the tumor type. However, HGG and LGG are difficult to differentiate, and misdiagnosis may lead to a suboptimal prognosis \cite{mzoughi2020deep}. 

% Among pediatric brain tumors, which are significantly different in genetic alterations from adult brain tumors, pediatric LGG (pLGG) is the most common central nervous system tumor in children and youth, accounting for more than 33\% of all pediatric brain tumors \cite{de2019management,krishnatry2016clinical}. Successful pLGG treatment planning should correctly identify the molecular subtype, which can identify key genetic events of pLGG \cite{pollack2019childhood,ryall2020integrated,sturm2017pediatric}. Although the prognosis of pLGG has an excellent overall survival rate between 85\%-96\% within five to ten years \cite{krishnatry2016clinical}, patients who survived from pLGG may suffer from the functional and neurological complications from the disease or treatment \cite{de2019management}. Hence, it is crucial to identify correct molecular subtypes before treatment. 

Among pediatric brain tumors, pediatric low-grade glioma (pLGG) stands out as the most prevalent central nervous system tumor in children and young individuals, constituting more than a third of all pediatric brain tumors \cite{de2019management,krishnatry2016clinical}. Successful planning of pLGG treatment relies on the accurate identification of its molecular subtype, and thus it is important to determine the key genetic events associated with pLGG \cite{pollack2019childhood,ryall2020integrated,sturm2017pediatric}. While the prognosis for pLGG is generally favorable, boasting an overall survival rate of 85\% to 96\% within a five- to ten-year timeframe \cite{krishnatry2016clinical}, survivors may still suffer from functional and neurological complications from the disease or its treatment \cite{de2019management}. Thus, it becomes imperative to determine the correct molecular subtypes before initiating treatment. Currently, the standard method for identifying pLGG molecular subtypes is through biopsy, which is invasive and there is a potential risk of infection or hemorrhage after biopsy \cite{namdar2022tumor}. To circumvent these drawbacks, medical imaging, especially MRI scans \cite{tak2023noninvasive}, emerges as a promising alternative to biopsy, enabling radiologists to diagnose specific molecular subtypes and facilitating more precise prognostic assessments. In this work, we focus on two common pLGG molecular subtypes BRAF Fusion and BRAF V600E Mutation.

In recent years, deep learning-based methods have proven to be one of the most powerful tools for pediatric and adult brain tumor classification tasks in both 2D and 3D MRI, and have provided promising performance with high accuracy and AUC (Area under the ROC Curve) \cite{ge2020deep,hao2021transfer,mzoughi2020deep,namdar2022tumor,pei2020brain,swati2019brain,tak2023noninvasive,tandel2020multiclass}. Deep learning approaches require a large amount of data to train, which is an ill-posed problem in medical imaging for rare diseases such as LGG and pLGG. This leads to poor performance of deep learning models since they tend to overfit the minority class data and cannot generalize well to an unseen dataset. There are several works that aim to mitigate the imbalanced data problem in recent years, which could be categorized into two main research areas. First, the transfer learning approach is a common way to address the problem by pretraining the model on a huge dataset (i.e., ImageNet that contains millions of natural images), and then fine-tuning the same model on a small, domain-specific dataset (i.e., brain tumor MRI data). Several works have shown the transfer learning approach can be beneficial in the context of medical data \cite{ashraf2021deep,ghazal2022alzheimer,srinivas2022deep,tak2023noninvasive,ullah2022effective} with promising performance. Another approach is to apply some data augmentation techniques, which can range from the most traditional image augmentations, e.g., flip, rotation, and translation \cite{mzoughi2020deep,rehman2020deep,sajjad2019multi} to the more advanced Generative Adversarial Network (GAN)- and Diffusion-based methods \cite{ge2020deep,kwon2019generation,volokitin2020modelling,khader2022medical,peng2022generating} that could also generate synthetic data to mitigate the need for large datasets. These GAN- and Diffusion-based methods focus on whole-image generation. However, a group of works \cite{cheng2015enhanced,pei2020brain,pereira2018automatic} have demonstrated the effectiveness of using the tumor region of interest (ROI) instead of the entire image slice or volume for classification because ROIs contain less redundant information and require less memory and computational resources when training the model. Therefore, augmenting ROIs is a more challenging task for GANs because the model has to either generate the image and the segmentation mask simultaneously \cite{subramaniam2022generating} or generate ROIs directly.

In this work, we propose a novel framework based on an auto-encoding GAN architecture that can generate high-resolution 3D MRI brain tumor ROIs using a small amount of data. More precisely, we extend the Vector-Quantized GAN (VQGAN) \cite{esser2021taming} to generate synthetic 3D brain tumor ROI of LGGs on the BraTS 2019 dataset and BRAF V600E Mutation on our internal pLGG dataset collected at The Hospital for Sick Children (SickKids), Toronto, Canada. The VQGAN model has the ability to generate high-resolution images while preventing mode collapse by using a combination of a convolutional neural network (CNN) and an auto-regressive transformer. We incorporate the masked token modeling strategy when training the transformer model. To the best of our knowledge, this work is the first attempt to solely generate brain tumor ROIs in 3D MRI that can be used as additional data for the imbalanced classification of brain tumor types. To validate the quality of generated tumor ROIs, we report various quantitative metrics and the performance of using the generated data for classifying tumor types on both datasets. Experiments show our proposed method outperforms baseline models in both objective image quality metrics and classification performance. Finally, we envision the proposed framework can be used in a wide range of medical imaging applications such as rare disease classification, as the method requires less data to train. In summary, the contributions of this work can be summarized as follows:

\begin{enumerate}
    \item We propose a novel 3D-VQGAN model with an auto-regressive transformer through a masked token modeling approach that can be applied to data-constrained unconditional image generation tasks such as tumor ROIs generation. This is also the first attempt to use masked modeling with the transformer for 3D brain tumor ROIs generation.
    \item We are the first to unconditionally synthesize LGG ROIs on BraTS dataset and BRAF V600E Mutation tumor ROIs on SickKids pLGG datasets in 3D MRI, and show that the generated data are more realistic than other GAN- or Diffusion-based architectures based on various image generation metrics.
    \item We show that the generated data from our proposed framework can be directly used as additional data in a downstream tumor type classification task on BraTS and SickKids pLGG datasets. The classification results further validate the superiority of the proposed method.
\end{enumerate}
 %than other deep-learning based data augmentation methods {\color{red}[give Ref]}.

%come with a life expectancy of several years so aggressive treatment is often delayed as long as
 
\section{Related Work}

Image generation has been revealed as an increasing research trend recently and has been shown promising performance for generating high-resolution 2D \cite{bao2021beit,chang2022maskgit,crowson2022vqgan,esser2021imagebart,esser2021taming,li2023mage,yu2021vector} and 3D data \cite{ge2022long,harvey2022flexible,ho2022imagen,yang2022diffusion}, as well as Magnetic Resonance (MR) images for brain studies \cite{khader2022medical,kwon2019generation,peng2022generating}. Early image generation methods fall into the GANs family, several GAN-based image generation frameworks have demonstrated the capacity of GANs model to synthesize realistic medical images through unconditional settings, e.g., generated from random noise \cite{han2018gan,hong20213d,kwon2019generation,volokitin2020modelling} or image-to-image translation, e.g., from other MR sequences \cite{dar2019image,shin2018medical,yu20183d}. However, both unconditional and image-to-image translation approaches have exhibited potential drawbacks. The former tends to produce blurry images and may encounter the mode collapse problem \cite{kwon2019generation}, while the latter needs a large number of well-curated data pairs, which is infeasible to obtain for some rare diseases. To address the potential limitations inherent to GAN-based models, Diffusion models have been proposed and have demonstrated superior performance over GANs \cite{muller2022diffusion}. Denoising Diffusion Probabilistic Model (DDPM) \cite{dhariwal2021diffusion,ho2020denoising} has shown impressive performance in medical image analysis by learning a Markov-chain process that transforms a sample Gaussian distribution to the target data distribution. Several works have been applied and extended DDPM to segmentation \cite{la2022anatomically}, anomaly detection \cite{wolleb2022diffusion}, and 3D MRI generation \cite{khader2022medical,peng2022generating}. However, DDPM is extremely computationally expensive, thus posing challenges in the training process. 

More recently, the autoregressive transformer has attracted great attention in the image generation tasks \cite{esser2021imagebart,esser2021taming,yu2021vector}. A pivotal technique integral to autoregressive image transformers is known as Vector Quantization Variational Autoencoders (VQ-VAE), which aims to learn images' low-dimensional features and represent them as discrete tokens \cite{razavi2019generating,van2017neural}. VQGAN \cite{esser2021taming} builds upon the VQ technique by incorporating adversarial and perceptual loss to enhance the perceptual quality of reconstructed images. ViT-VQGAN \cite{yu2021vector} further improves the VQGAN framework by replacing the conventional CNN encoder and decoder with the Vision transformer (ViT) counterparts. The application of the autoregressive transformer has also extended to the domain of medical images \cite{pinaya2023generative,tudosiu2022morphology}, both works aim to generate high-resolution brain MR images using the VQ-VAE to learn image discrete representations and the subsequent transformer to learn the underlying data distributions. Nevertheless, it is noteworthy that both methodologies are constrained to modeling major or minor pathological areas within the brain, which is the typical scenarios encountered in real clinical settings. The transformer-based image generation for brain MR images with pathological areas have not been well explored, which becomes a critical bottleneck for deploying the model to clinical routines.  
% However, both methods are limited to modeling major or minor pathological areas in the brain, which is the common scenario in the real clinical setting.

Masked modeling has emerged as a prominent technique when training the transformer model for image generation tasks. ViT explores the masked patch prediction for self-supervision. Building upon the foundations of BERT \cite{devlin2018bert}, BEiT \cite{bao2021beit} extends this concept to directly predict visual tokens within images. Recently, several studies have applied masked modeling for VQ-based image generation. MaskGiT \cite{chang2022maskgit} introduces an innovative approach, combining a mask scheduling strategy with parallel decoding to synthesize images effectively using bidirectional transformers. Meanwhile, MQ-VAE \cite{huang2023not} introduces a unique masked VQ-VAE, where unimportant region features are masked during image representation learning. This model employs a stackformer to autoregressively predict the next visual token and its respective position. To bridge the gap between medical image generation and masked modeling approach, we propose a two-stage VQGAN and transformer that incorporates masked modeling in this work. It is also worth to highlight the differences between our proposed method and two similar works above \cite{pinaya2023generative,tudosiu2022morphology}: (1) Our primary focus lies in generating brain tumor ROIs rather than healthy or whole-brain images. This task is particularly challenging due to the small size of ROIs and the inherent difficulty in capturing their unique features. Moreover, the limited availability of data for ROIs further complicates the task. (2) We use the masked modeling technique for training the transformer model to learn the underlying distributions. Our contribution centers on advancing 3D medical image generation through the masked modeling approach. To the best of our knowledge, this is the first time the masked approach within the transformer has been applied to generate brain tumor ROIs.

\section{Materials and Methods}

% In this section, we start by talking about the two-stage framework, 3D-VQGAN and Transformer models in detail, and then talk about the two datasets we utilized in this work.

\subsection{Model Architecture} \label{model_arc}
There are two main challenges for generating 3D images using GANs. The first one is mode collapse where the generator produces limited variations of images, and the second is that conventional GANs tend to produce low-resolution images, especially for smaller tumor ROIs. To address the above problems, we propose 3D-VQGAN, a solution based on VQGAN \cite{esser2021taming} that generates high-resolution images by incorporating a CNN-based autoencoder to extract the local features and learns a codebook for representing the context of the image, and use another subsequent transformer model to learn the long-term interrelations of the image compositions. The core of the VQGAN model is the vectorized latent representations and a corresponding codebook. More precisely, the latent feature maps in the bottleneck of the autoencoder are mapped to the quantized representation, i.e., a sequence of semantic tokens, from a learned codebook. The codebook is the key to generating high-resolution images \cite{huang2023not}. More details are in the following section. For generating 3D tumor ROIs, we follow the original VQGAN architecture, and we use 3D convolutional layers instead of 2D to extract not only the spatial information but also the 3D context in the latent space. The overall framework is trained in two stages. First, we train the encoder, codebook, decoder, and discriminator to learn the efficient data representation through a reconstruction task. Second, we freeze the modules trained in the previous stage and only train the transformer that aims to auto-regressively predict the next semantic token in the quantized representation.

\begin{figure}[t]
     \centering
     %\begin{subfigure}[b]{0.32\textwidth}
     \centering
     \includegraphics[width=0.95\textwidth]{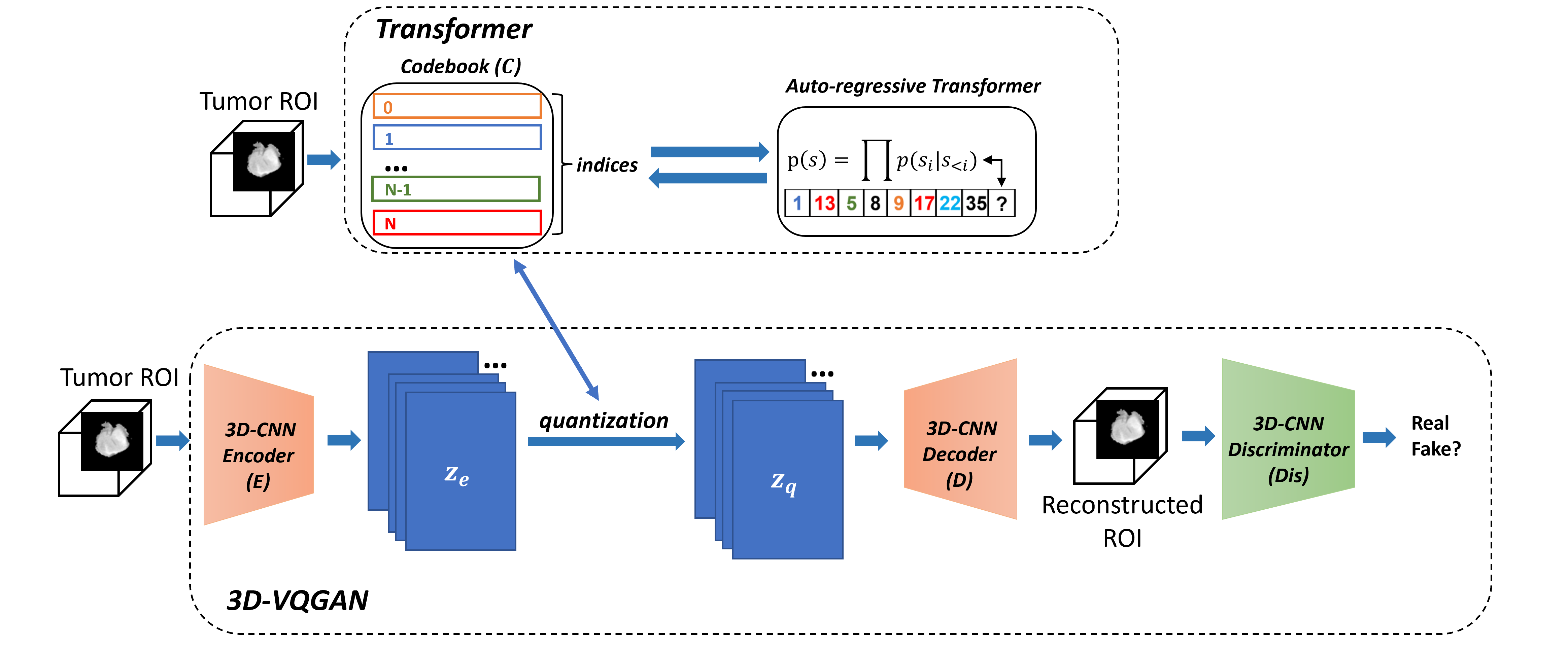}
     \caption{Detailed overview of the proposed method. Our method contains two modules, 3D-VQGAN and the auto-regressive transformer. The former maps both the spatial and depth information of 3D tumor ROIs into discrete tokens. The latter uses the transformer architecture to capture the long-term dependency.}
     \label{model}
\end{figure}

\textbf{Stage 1. 3D-VQGAN:} The first stage follows the GAN training paradigm as shown in the bottom of Figure \ref{model}. Recall that the codebook is trained in this stage as well. Our encoder $E$ and decoder $D$ consist of five 3D convolution layers and five 3D residual blocks. In $D$, we use the upsample-convolution approach instead of transpose convolution to prevent the checkerboard artifacts. For all residual blocks and convolution layers in $D$, we use a kernel size of $3 \times 3 \times 3$, followed by the Rectified Linear Unit (ReLU) activation function. The kernel size of the convolution layers in $E$ is $4 \times 4 \times 4$, followed by the Leaky ReLU activation function. Following \cite{esser2021taming}, we use a discriminator $Dis$ to differentiate the original and reconstructed image. Our discriminator has five 3D convolution layers with kernel size $4 \times 4 \times 4$ followed by Leaky ReLU activation functions. Different from \cite{esser2021taming}, we remove the attention block in both the encoder and the decoder to lower the computational resources and use the 3D image gradient operator to model the fine details instead. We use batch normalization to stabilize the training process, and the $tanh()$ function is placed over the outputs of $D$ to produce the final image. In the quantization step, the latent feature maps are quantized by replacing each one with its closest corresponding codebook vector in the learned codebook $C$. Formally, we train a learnable codebook $C = \{c_i\}_{i=1}^{K}$ that transforms feature vectors $z_e \in \mathbb{R}^{H\times W \times Dp \times n_z}$ encoded from $E$ to $K:=H\times W \times Dp$ discrete latent tokens $c_q, q \in [1,K]$ by the nearest neighbor search method in $C$, and each token $c_q$ includes an embedding vector $c_z \in \mathbb{R}^{n_z}$. Finally, we stack $K$ quantized feature vectors $c_{\text{stack}} \in \mathbb{R}^{H\times W \times Dp \times n_z}$ and feed it into the decoder $D$ to produce reconstructed images. We denote $H$ as the height, $W$ as the width, $Dp$ as the depth, $n_z$ as the number of feature maps, and $K$ as the total number of discrete tokens in $C$.

\begin{figure}[t]
     \centering
     %\begin{subfigure}[b]{0.32\textwidth}
     \centering
     \includegraphics[width=0.95\textwidth]{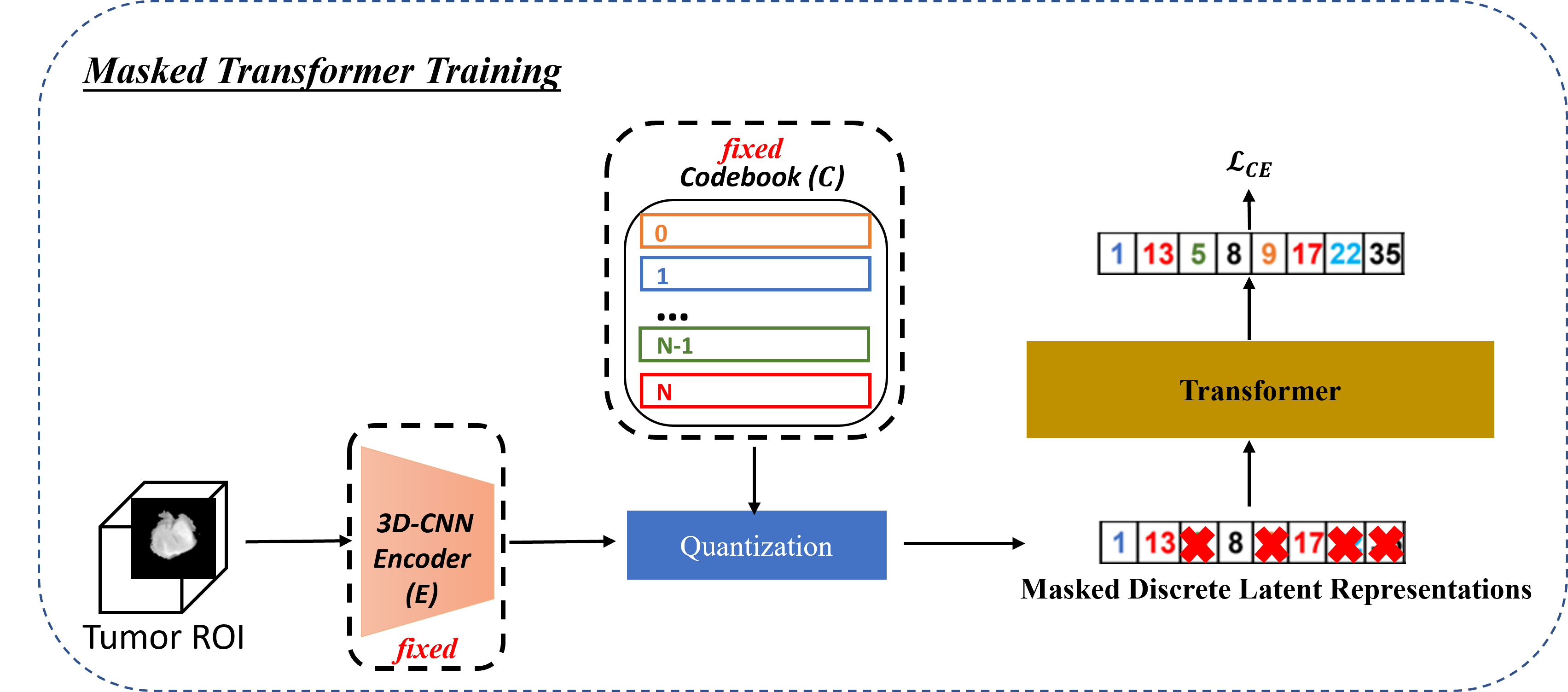}
     \caption{Detailed overview of the training phase for transformer model. We fix the Encoder $E$ and the Codebook $C$, and only trained the transformer model.}
     \label{model_stage2}
\end{figure}

\textbf{Stage 2. Masked Token Modeling with Transformer:} The masking strategy is first introduced in BERT \cite{devlin2018bert} designed for the masked language modeling task for natural language understanding and representation learning. In computer vision, the same masking approach has been extended to image representation learning tasks by directly masking on the image pixels \cite{bao2021beit,he2022masked} or on the discretized tokens after image quantization \cite{chang2022maskgit,li2023mage}. In this work, we adapt the masking strategy on discrete semantic tokens obtained by the codebook $C$ in the first stage. Specifically, we have $E, D, C$ trained in this stage and we freeze these modules, hence we can represent 3D images in the latent space and further quantize to indices $\mathbf{c}$ in $C$ and their corresponding embeddings. Formally, let $\mathbf{Y} = \{y_i\}_{i=1}^{K}$ be the linearized discrete tokens obtained from our codebook $C$ trained in the first stage. We use the raster-scan order to perform the linearization. Let $\mathbf{M} = \{m_i\}_{i=1}^{K}$ be the mask for each of the discrete tokens, where $m_i = 1$ if the token $i$ is \textit{unmasked} and $m_i = 0$ if the token $i$ is \textit{masked out}. For those masked-out tokens, we adapt the BERT masking strategy \cite{devlin2018bert} to replace them with the random indices in the codebook $C$. Then, the training objective is to reconstruct the masked tokens using unmasked tokens. Our hypothesis is that by masking out some of the tokens, the transformer can better learn the relationships between the semantic tokens and improve its learning ability. The transformer can model the prior categorical distribution of $\mathbf{Y}$, which is sequentially given by $p(y_0)\prod_{i=1}^{K} p(z_{i+1}|z_{\leq i})$ where $y_0$ represents the start of sequence token. Since we are masking out tokens, we can use the cross entropy between the reconstructed token sequence and the ground truth token sequence. The loss function is detailed in the next section. A graphical illustration of the training process for the masked transformer model is depicted in Figure \ref{model_stage2}.

%In the second stage, we have $E, D, C$ trained, and thus we can represent 3D images in the latent space and further quantize the latent space to indices $\mathbf{c}$ in $C$ and their corresponding embeddings. For the transformer model in this stage, we use the same GPT-2 model, as done in \cite{esser2021taming} and treat it in an auto-regressive manner. To train the transformer, we use a ``masked token modeling'' approach, which shares a similar idea in masked image modeling \cite{xie2022simmim}, where we randomly mask out half of the total quantized indices, and let the transformer predict and recover masked indices. %as output $\hat{z}$.
%The transformer can be used to model the prior distribution of $\mathbf{c}$, and we can further use the negative log-likelihood as the loss to train the model.

For the downstream classification task between different tumor types, we use a standard 3D ResNet-50 model \cite{hara2017learning} that takes 3D tumor ROIs as inputs and outputs two class probabilities that indicate which tumor type the inputs belong to.

%There are two loss functions used for training the whole VQGAN pipeline.
\subsection{Loss Function}

In the first stage of the overall training pipeline, we use the combination of the pixel differences loss ($\mathcal{L}_1$), perceptual loss ($\mathcal{L}_{perp}$) \cite{johnson2016perceptual}, GAN-based feature matching loss ($\mathcal{L}_{match}$) \cite{ge2022long}, 3D image gradient loss ($\mathcal{L}_{grad}$), codebook loss ($\mathcal{L}_{codebook}$) \cite{esser2021taming}, and the discriminator loss ($\mathcal{L}_{Dis}$). Given the original image as $x$ and the reconstructed image as $\hat{x}$, the pixel differences loss, perceptual loss \cite{johnson2016perceptual}, and the GAN-based feature matching loss are shown in Equation \eqref{l1perpmatch}. For $\mathcal{L}_{perp}$, $f^{i}(x_j)$ is the feature map for the $j$-th random slice of $x$ and $\hat{x}$ in the $i$-th layer of the pre-trained VGG16 network \cite{johnson2016perceptual}, which is used as a feature extractor to extract and model the deep semantic similarly between feature maps. $\mathcal{L}_{match}$ is used to help stabilize the training process, where $f_{Dis}^{i}$ is the $i$-th layer of the trained discriminator.

\begin{equation} \label{l1perpmatch}
    \mathcal{L}_1 = \|x - \hat{x}\|_1 ,\ 
    \mathcal{L}_{perp} = \sum_{j=1}^{3}\|f^{i}(x_{j}) - f^{i}(\hat{x_j})\|_{2}^{2} ,\ 
    \mathcal{L}_{match} = \|f_{Dis}^{i}(x) - f_{Dis}^{i}(\hat{x})\|_{1}
\end{equation}

\textbf{Impact on using the image gradient loss: }The image gradient loss is used to learn the fine-grained information in 3D medical images. Our hypothesis is that the gradient loss will encourage the reconstructed image to preserve as many details as in the original images, such as anatomical consistency and tissue correctness. Since the latent features are first been quantized and replaced by embedding vectors from the learned codebook $C$, the gradient loss will also enforce the latent features before and after quantization to contain as many details as possible in the feature space. Inspired by previous works \cite{mathieu2015deep,sanchez2018brain}, we take advantage of the Axial (A), Sagittal (S), and Coronal (R) planes in 3D MRI images to design the gradient loss. Thus, the 3D image gradient loss is proposed as shown in Equation \eqref{grad}:

\begin{align} \label{grad}
    \begin{split}
        \mathcal{L}_{grad} = \|\nabla(A(x))-\nabla(A(\hat{x}))\|_2^2+\|\nabla(R(x))-\nabla(R(\hat{x}))\|_2^2+\|\nabla(S(x))-\nabla(S(\hat{x}))\|_2^2
        % \mathcal{L}_{grad}^{C} = \|\nabla(x[:,i,:]) - \nabla(\hat{x}[:,i,:])\|_2^2 \\
        % \mathcal{L}_{grad}^{S} = \|\nabla(x[i,:,:]) - \nabla(\hat{x}[i,:,:])\|_2^2
	\end{split}
\end{align}

\noindent $\nabla(\cdot)$ computes the $x$- and $y$-direction gradients of the image. $A(x) = x[:,:,i]$ represents slicing over axial plane for $i$-th slice, and similarly, $R(x) = x[:,i,:]$ and $S(x) = [i,:,:]$ for slicing over coronal and sagittal plane, respectively. The discriminator loss, shown in Equation \eqref{hinge_dis}, aims to differentiate between the real and the reconstructed image. We use the hinge-loss variant instead of the vanilla loss. We follow the same codebook loss described in \cite{esser2021taming}, also shown in Equation \eqref{codebook_loss}, where $sg[\cdot]$ represents the stop gradient operation. The gradient can not be back-propagated through the network because the quantization operation is not differentiable \cite{esser2021taming}, and hence we use an EMA update as in \cite{ge2022long} to optimize the first loss term of $\mathcal{L}_{codebook}$.  

\begin{equation} \label{hinge_dis}
    \mathcal{L}_{Dis} = \mathbb{E}_{x\sim p_{d}}[max(0,1-D(x))] + \mathbb{E}_{\hat{x}\sim p_{\hat{d}}}[max(0,1+D(\hat{x})]
\end{equation}

\begin{equation} \label{codebook_loss}
    \mathcal{L}_{codebook} = \|sg[E(x)]-c_z\|_2^2 + 0.25\|sg[c_z]-E(x)\|_2^2
\end{equation}

Aggregating all the loss terms together yields the final loss objective in Equation \eqref{loss_obj} for the first stage of the framework:

% \begin{align}
%     \begin{split}
%         \min_{E,D,C}(\max_{Dis}(\mathcal{L}_{Dis})) \\
%         \min_{E,D,C}(\mathcal{L}_{1} + \mathcal{L}_{perp} + \mathcal{L}_{match} + \mathcal{L}_{grad} + \mathcal{L}_{codebook})
%     \end{split}
% \end{align}

\begin{equation}
    \begin{gathered}
    \min_{E,D,C}(\max_{Dis}(\mathcal{L}_{Dis})) \\
    \min_{E,D,C}(\lambda_1\mathcal{L}_{1} + \lambda_2\mathcal{L}_{perp} + \lambda_3\mathcal{L}_{match} + \lambda_4\mathcal{L}_{grad} + \lambda_5\mathcal{L}_{codebook}) \label{loss_obj}
    \end{gathered}
\end{equation}
% \begin{gather*}
%     \min_{E,D,C}(\max_{Dis}(\mathcal{L}_{Dis})) \\
%     \min_{E,D,C}(4 * \mathcal{L}_{1} + \mathcal{L}_{perp} + 4 * \mathcal{L}_{match} + 4 * \mathcal{L}_{grad} + \mathcal{L}_{codebook})
% \end{gather*}

Where $\lambda_i, i\in[1,5]$ is the weighting factor between different loss terms. We follow previous works \cite{ge2022long,khader2022medical} to set $\lambda_1 = \lambda_3 = 4$ and $\lambda_2 = \lambda_5 = 1$. Since the image gradient loss is as important as the pixel $\mathcal{L}_1$ loss, we set $\lambda_4 = 4$ as well. 

For the transformer model as described in Section \ref{model_arc}, we use the cross entropy loss between the reconstructed token sequence and the ground truth token sequence as shown in Equation \eqref{masked_loss} to optimize the transformer. 
%For the second stage of the proposed framework, we follow exactly the same procedure as in \cite{esser2021taming}: 

\begin{align} \label{masked_loss}
    %\mathcal{L}_{transformer} = \mathbb{E}_{z\sim p(z_{data})}[-\log p(z)]
    \mathcal{L}_{transformer} = -\mathbb{E}_{\mathbf{Y}\in \mathcal{D}}(\sum_{\forall i, m_i = 0}log p(y_i|\mathbf{Y}_{M}))
\end{align}

Where $\mathcal{D}$ is the training dataset, $\mathbf{Y}_{M}$ denotes the \textit{unmasked} tokens, thus the masked tokens can conditioned on these unmasked tokens during training.
%\noindent Where $z$ is the discrete representation of the input images. $p(z)$ is sequentially given by $p(z_0)\prod_{i=1}^{h \times w \times c} p(z_{i+1}|z_{\leq i})$ from the autoregressive transformer model and $z_0$ represents the start of sequence token.

\subsection{Data and Preprocessing}

\begin{figure}[ht]
     \centering
     %\begin{subfigure}[b]{0.32\textwidth}
     \centering
     \includegraphics[width=\textwidth]{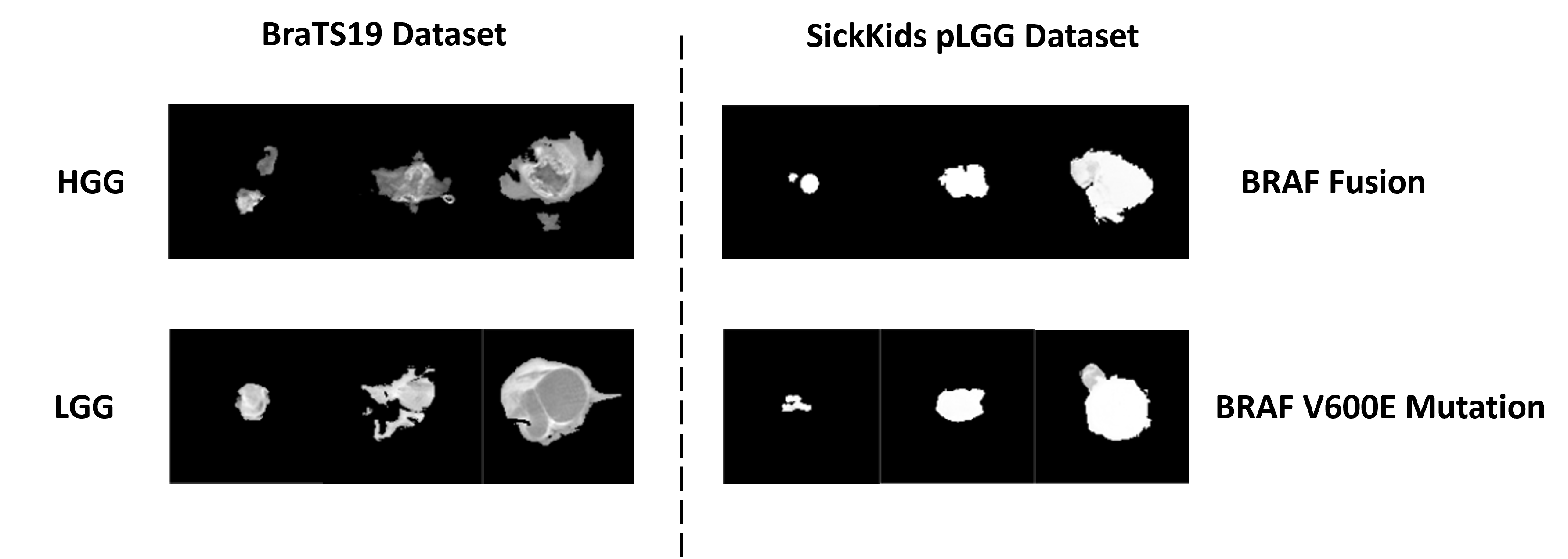}
     \caption{A sample visualization of the data in both datasets. We show the center slice in the axial plane of the MRI data for three different samples for different tumor types.}
     \label{sample_dat}
\end{figure}

We utilized two datasets in this work to demonstrate the superior performance and robustness of the proposed method. First, we used the FLAIR sequence of the MRI data from the publicly available BraTS 2019 dataset \cite{bakas2017advancing,bakas2018identifying,menze2014multimodal} contains 335 patients ($n_{b}$=335), with 259 HGG patients and 76 LGG patients \footnote{ https://www.med.upenn.edu/cbica/brats2019/data.html}. Second, we used the FLAIR sequence of the MRI data from our internal SickKids pLGG dataset ($n_{p}$=214) containing 143 patients with BRAF Fusion and 71 patients with BRAF-V600E Mutation. For both datasets, the original whole-brain images and corresponding segmentation masks are provided either by the dataset provider or experienced neuro-radiologists. For both datasets, we reshape the data from $240 \times 240 \times 155$ to a uniform shape of $128 \times 128 \times 128$. To achieve this, we first remove all zero-valued slices in both the whole brain images and the segmentations, since we are interested in the slices with the brain tumor present. Then, we obtain the ROIs by multiplying the whole-brain images with the segmentation masks. Finally, we center crop (or pad if the number of slices is already less than 128) the region based on the segmentation mask to a target size of $128 \times 128 \times 128$. A visualization of sample images from both datasets is presented in Figure \ref{sample_dat}.

\section{Experiments} \label{exp_res}

% {\color{red} need pLGG quantitative and qualitative results, e.g., mmd, ms-ssim, fid}
% In this section, the details of the experiment are presented. We also provide the visual results of the generated ROIs from the proposed method, along with two relevant qualitative metrics. Finally, the performance of the classification task, trained with the generated ROIs is provided.
In this section, we provide the experimental details on both BraTS and pLGG datasets. All programs were implemented in Pytorch, and all models were trained on a single 16 or 32GB TESLA V100 GPU depending on which one was available in the cluster. Additionally, we applied the automatic mixed precision in the PyTorch library during the training process \cite{subramaniam2022generating} to alleviate the computation costs caused by 3D inputs. We conducted the main experiments on the BraTS 2019 dataset and further validated the effectiveness and robustness of the proposed method in different brain tumor types from the SickKids pLGG dataset.

\subsection{Experiments Details for BraTS Dataset} \label{exp_det_brats}
For the first stage of the proposed model, we train for 4000 epochs with an initial learning rate of 0.0001 and cosine decay to 0 for all sub-modules, a mini-batch size of 3, and the Adam optimizer. We set the codebook size $K = 512$. For the second stage, we train the transformer for 1500 epochs using a learning rate of $4.5e-06$, a mini-batch size of 3, and the AdamW optimizer \cite{loshchilov2017decoupled}. We mask out half of the total discrete tokens (mask ratio = 0.5) when training as described in Section \ref{model_arc}. The total training process takes about 36 hours to complete. We term the model with latent space dimension, i.e. the output from Encoder $E$, with size $4 \times 4 \times$ 4 (resp. $8 \times 8 \times 8$) as 3D-VQGAN-lat4 (resp. 3D-VQGAN-lat8). We normalize all images within the range of $[-1, 1]$ to match the output from decoder $D$. We randomly hold out 25 patients from both HGG and LGG as a standalone test set; these data are hidden from either training the 3D-VQGAN or the classifier. For the remaining 51 LGG patients, we use all to train the 3D-VQGAN model. The rest of the 234 HGG patients are only used for our classification task.

% {\color{red} \textbf{Why make balance test/train on HGG and LGG and why low performance for baseline} low performance because of balanced train test on HGG and LGG. Since our ultimate goal is to make the training set balance by using synthetic data, we want to access the actual performance of the minority class, then we select a third of the original data as test set. Also, because except baseline model, every other models are trained with equal amount of HGG and LGG, so this applies to test set as well, we have the same number of amount of HGG and LGG. The baseline here is just a clinical reference, baseline+trad.aug. should be the real baseline}

For classification, we design three sets of training data combinations: $(a)$ 234 HGG, 51 LGG, $(b)$ 234 HGG, 234 LGG, where we apply traditional augmentation techniques, such as rotation by 30 degrees, scaling by 1.5 times larger, left-right flipping, and elastic deformation to form a balanced dataset, and $(c)$ we first pretrained a classifier with 183 real HGG and 183 synthetic LGG generated from either our model or other baseline models, and then finetuned with 51 real HGG and 51 real LGG, following the setup in \cite{khader2022medical}. We use the 3D ResNet-50 model \cite{hara2017learning} and trained for 50 epochs for all classification experiments. Models for combination $(a)$ and $(b)$ are trained with a batch size of 8 and a learning rate of 0.01. For $(c)$, we use a batch size of 8, a learning rate of 0.001 with Adam optimizer for pretraining, and then finetuned using a batch size of 10, and a learning rate of 0.1. All classification models use focal loss \cite{lin2017focal} with or without explicit class weights depending on the number of data in each class. The class weights are computed as the number of samples in the target class divided by the total training samples.

In addition, we have chosen the model trained with training data combination $(a)$ as the reference model for the following reasons: The primary objective of our work is to augment minority class data, thereby creating a balanced dataset for training. To ensure consistency between training and testing, both datasets should be balanced. However, training data combination $(a)$ consists of an \textit{imbalanced} training set and a \textit{balanced} test set. We have included this model because it closely aligns with real-life scenarios, illustrates the performance when no augmentations are applied, and further infers the superiority of our proposed method. Furthermore, our decision to employ a balanced test dataset is two-fold. Firstly, it maintains consistency with the training dataset, as we mentioned previously. Secondly, it yields more accurate performance for predicting minority class data. If we maintained the original class ratio in the test dataset, the results might be heavily influenced by the data in the majority class.
% We use the balanced dataset not only because it is consistent with the training dataset, but also because it can provide more accurate performance in predicting the minority class data. If we maintain the ratio in the original training dataset, the results might be dominated by the majority class data. 

Due to computational limitations, doing cross-validation throughout the training of the proposed model is hard. It would be expensive and time-consuming given our current computational resources. However, we ensure robustness and reliability in our findings by performing multiple trials for all classification experiments. Specifically, we use 85\% of the training data for optimizing the model and the remaining 15\% of the training data for validation, and we repeat this process three times. In the validation set, we still maintain the balanced ratio between two classes. We also ensure that there is no overlap between the validation data in the three runs. We believe that our current approach can still provide valuable insights into the performance of our proposed method.

\noindent \textbf{Baseline Model. } For comparison, we use the 3D-WGAN-GP \cite{gulrajani2017improved}, 3D-$\alpha$WGAN \cite{kwon2019generation}, which is one of the state-of-the-art GAN-based methods designed for the whole brain generation, and a diffusion-based Medical Diffusion \cite{khader2022medical} as our baselines for tumor ROIs generation. We reimplement and rerun all baselines. We use them to demonstrate the superior performance of our proposed method when only tumor ROIs are provided. Both 3D-WGAN-GP and 3D-$\alpha$WGAN models are trained for 2000 epochs. The input to 3D-$\alpha$WGAN is a 1000-dimensional random vector $\mathbf{z_{a}}$ as suggested in \cite{kwon2019generation}, whereas the input to 3D-WGAN-GP is a 128-dimensional random vector $\mathbf{z_{b}}$. For Medical Diffusion, we follow the exact settings in the original paper \cite{khader2022medical}, except we only train 10000 epochs for the diffusion model due to computational limitations. We establish a baseline for the classification task where we use traditional augmentations, data combination $(b)$, for a fair comparison with GAN- and Diffusion-based augmentation methods. We also provide the results for our reference model mentioned above. 

% For the first baseline, we use data combination $(a)$ which best reflects the real-life scenarios. For the second baseline, we use data combination $(b)$ for a fair comparison with GAN-based augmentation methods. 
% Given that there are no papers that have employed a procedure similar to the one proposed in this work, we establish our own baseline for the classification task. For the first baseline, we use training data combination $(a)$ because this best reflects the real-life scenarios; an imbalanced data with a limited number of patients who has LGG. For the second baseline, we use training data combination $(b)$.

\subsection{Experimental Details for pLGG Dataset} \label{exp_det_plgg}

The parameters and training settings for 3D-VQGAN and the transformer model are the same as we described in Section \ref{exp_det_brats}, except the learning rate for the 3D-VQGAN model is set to be 0.00005 and the mask ratio is 0.15. We randomly hold out 20 patients from both BRAF Fusion and BRAF V600E Mutation molecular subtypes as a standalone test set; these data are hidden from either training the 3D-VQGAN or the classifier. For the rest of the 51 BRAF V600E Mutation patients, we use all patients to train the 3D-VQGAN model. The rest of the 123 BRAF Fusion patients are only used for our classification task.

For classification, we exactly follow Section \ref{exp_det_brats}, but now the training data combination $(a)$ contains 123 BRAF Fusion and 51 BRAF V600E Mutation; $(b)$ contains 123 BRAF Fusion and 123 BRAF V600E Mutation using traditional augmentation techniques, such as rotation by 30 degrees and left-right flipping to form a balanced dataset; and $(c)$ we first pretrained a classifier with 72 real BRAF Fusion and 72 synthetic BRAF V600E Mutation data generated from either our model or other baseline models, and then finetuned with 51 real patients from both subtypes, following the setup in \cite{khader2022medical}. The classification settings are the same as those for BraTS in Section \ref{exp_det_brats}. The reference and baseline models are also the same as described in the previous section.

\subsection{Generating Synthetic MRI Data}

To generate 3D tumor ROIs, we select a random index to start with from the codebook, and then have the transformer model predict and complete the rest of the indices for the 3D-VQGAN model. Once we have full indices, we get the corresponding embedding vectors for each of the indices and then feed them into decoder $D$ to produce the final images. For the baseline 3D-WGAN-GP and 3D-$\alpha$WGAN model, we sample the random noise vector $\mathbf{z_{b}}, \mathbf{z_{a}} \sim \mathcal{N}(0,1)$, respectively, and then fed it to the generator to produce the final images. For the Medical Diffusion model, we follow the same setup in \cite{khader2022medical} to generate the latent features of size $16 \times 16 \times 16$ from the diffusion model and decode to the output image by the VQGAN decoder $D$. All generated images are with size $128 \times 128 \times 128$.

\section{Results and Discussions} \label{qual_analysis}

In this section, we start with the qualitative and quantitative results on the BraTS dataset and give a thorough analysis and discussion, then we move to the results on our local pLGG dataset.

\subsection{Results for Generated Images on BraTS Dataset} \label{qua_ana_brats}

\begin{figure}[t]
     \centering
     %\begin{subfigure}[b]{0.32\textwidth}
     \centering
     \includegraphics[width=1\textwidth]{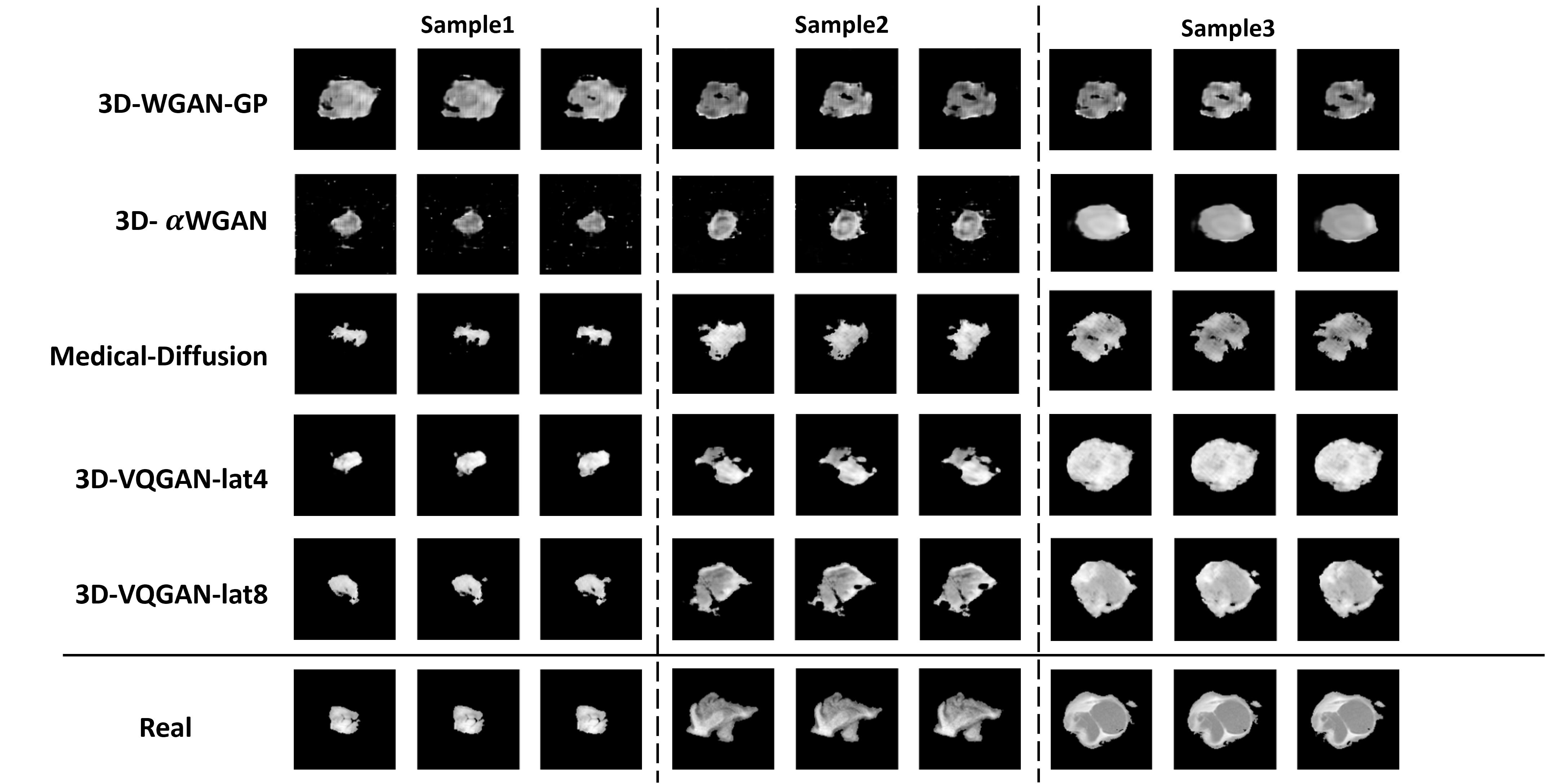}
     \caption{Qualitative comparison between generated and real LGG ROIs of three sample tumors that have different sizes. We show the center three slices in the axial plane for each ROI data. The suffix ``lat4'' represents the latent feature map dimension is $4 \times 4 \times 4$, and the ``lat8'' represents the latent feature map of size $8 \times 8 \times 8$.}
     \label{qual_res}
\end{figure}

\begin{figure}[t]
     \centering
     %\begin{subfigure}[b]{0.32\textwidth}
     \centering
     \includegraphics[width=1\textwidth]{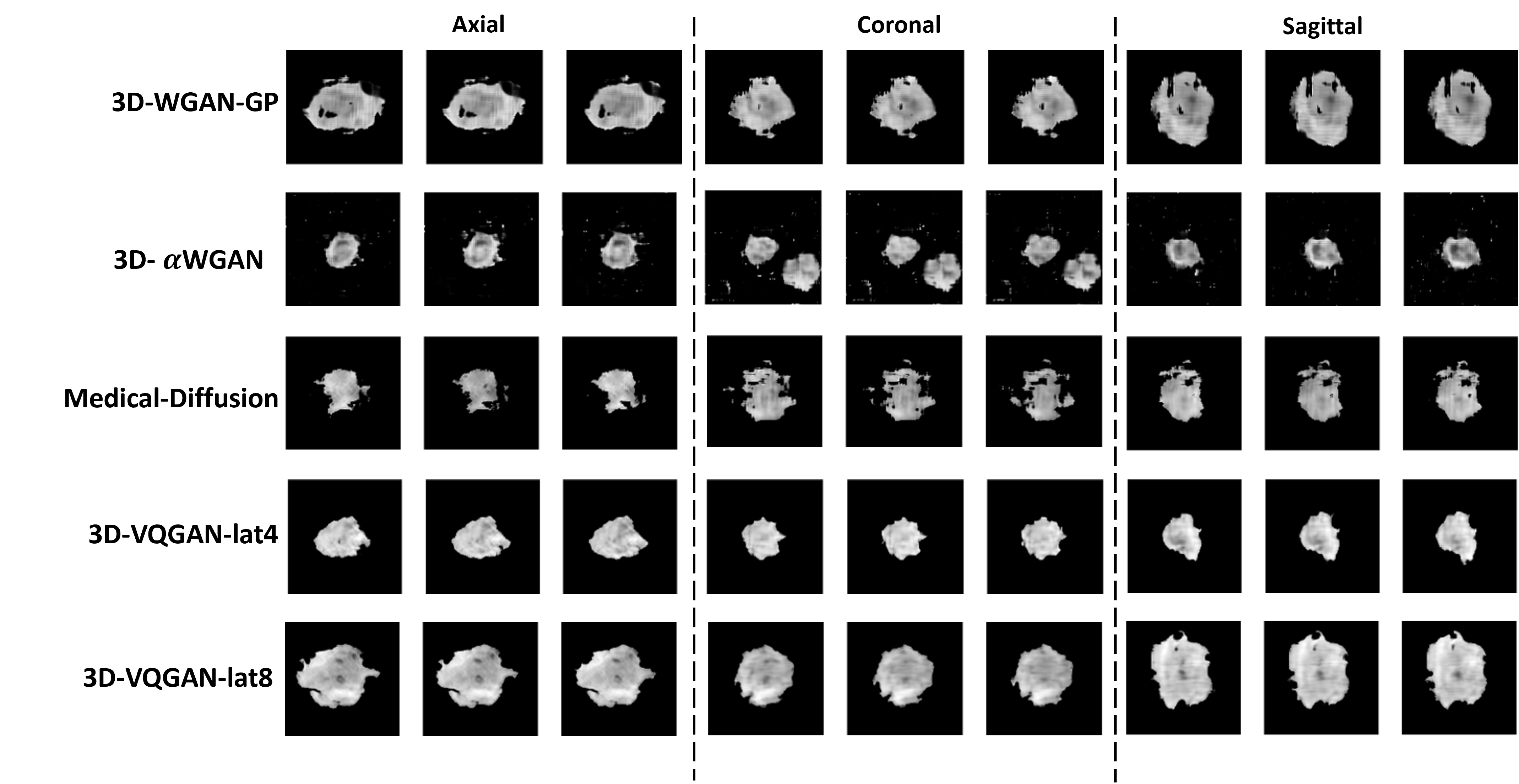}
     \caption{Qualitative comparison between generated LGG ROIs from our proposed model and other baseline models visualized in the axial, coronal, and sagittal plane. We show the center three slices. Figures have been enlarged to show the high fidelity of synthetic ROIs generated by our approach.}
     \label{qual_res_acs}
\end{figure}

In Figure \ref{qual_res}, we compare three different generated LGG ROIs from the baseline models and our proposed method with real LGG ROIs. The center three slices in the axial plane are shown for better visual quality. We also provide another visualization that focuses on Axial, Coronal, and Sagittal directions for a generated sample as these three planes are important in the context of medical images, shown in Figure \ref{qual_res_acs}. Notice that the generated sample selected for all models in Figure \ref{qual_res_acs} is different from the samples in Figure \ref{qual_res}. 

\begin{figure}[t]
     \centering
     %\begin{subfigure}[b]{0.32\textwidth}
     \centering
     \includegraphics[width=1\textwidth]{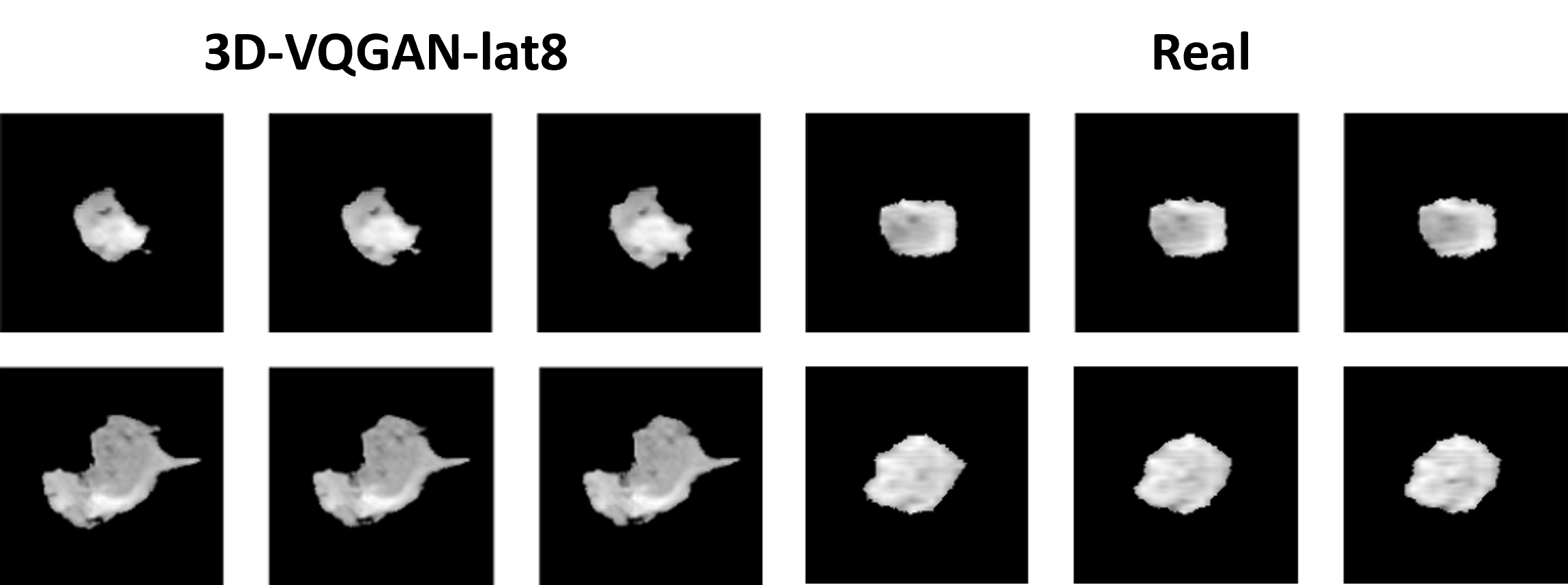}
     \caption{An example of LGG ROIs generated by our model and their closest real ROIs based on 3D-SSIM score. We enlarged the figure and only show the center three slices in the Axial plane.} %Top row 3D-SSIM Score between 3D-VQGAN-lat8 and Real: 0.93; Bottom row 3D-SSIM Score between 3D-VQGAN-lat8 and Real: 0.90.
     \label{qual_res_mostlike}
\end{figure}

Looking closely at the generated samples, both baseline models 3D-WGAN-GP and 3D-$\alpha$WGAN produce images with a lack of details, unexpected artifacts, and low image fidelity compared to the real ROIs. For the Medical Diffusion baseline, the generated images are better than the previous two baseline models, however, they still suffer from the minor checkerboard artifacts produced by the model. In contrast, generated samples from our proposed framework contain the detailed attributes of the tumor and exhibit high image fidelity. To quantitatively evaluate the image quality of the generated samples, we follow previous works \cite{khader2022medical,kwon2019generation,peng2022generating,tudosiu2022morphology,volokitin2020modelling} to compute the following three metrics: maximum mean discrepancy (MMD) score \cite{gretton2012kernel}, multi-slice structure similarity (MS-SSIM) score \cite{rosca2017variational}, and the Fr\'echet Inception Distance (FID) \cite{heusel2017gans}. We randomly selected 153 generated samples (three times larger than the training data size) to compute the above three metrics. MMD measures the distance between the generated and the real distributions where a lower value indicates more fidelity to the real data distribution. Due to memory constraints, we compute the batch-wise MMD$^2$ for the entire data over 100 tests with a batch size ($B$) of 3 and report the average score. Table \ref{quan_res} shows our 3D-VQGAN-lat8 model results in a lower MMD score, indicating that the distribution of our generated samples is the closest to the real distribution. The MS-SSIM score measures the diversity of generated samples, which computes the pairwise similarity over those samples. The closest score to the Real indicates the best. Here, we compute 1000 randomly sampled pairs and report the average score. Our 3D-VQGAN-lat4 model seems to struggle with generating diverse ROIs, this might be a sign of having the mode collapse problem. Nevertheless, our proposed 3D-VQGAN-lat8 model results in the lowest MS-SSIM, indicating the diversity is preserved. We also notice that 3D-$\alpha$WGAN fails to escape the mode collapse problem when dealing with small tumor ROIs both quantitatively from Table \ref{quan_res} and qualitatively from Figure \ref{qual_res}. Finally, the FID is also a common image generation metric for comparing the distribution between real and generated samples. We compute the FID in three views (e.g. Axial, Coronal, and Sagittal) to better reflect the nature of medical images. In Table \ref{quan_res}, we can see that our 3D-VQGAN-lat8 model results in the lowest FID in all three views, which coalesces with our previous analysis in other metrics that the 3D-VQGAN-lat8 model indeed produces high-fidelity and diverse images that looks very similar to real images. To examine the generated samples are not exactly the copies of training data, we give an example of two randomly selected samples and their closest real samples from the training cohort based on the 3D-SSIM \cite{sanchez2018brain} score in Figure \ref{qual_res_mostlike}. We can see that the generated samples are different from the real data.

\begin{table}
\centering
\caption{Quantitative results of generated images. Lower values indicate better performance for all metrics except MS-SSIM. For MS-SSIM, having the closest value to the Real represents the best. FID-A, -C, -S refers to calculating the FID over the Axial, Coronal, and Sagittal plane, respectively}
\begin{tabular}{c|c|c|c|c|c}
           & MMD(B=3) $\downarrow$    & MS-SSIM & FID-A $\downarrow$ & FID-C $\downarrow$ & FID-S $\downarrow$  \\ \hline
3D-WGAN-GP \cite{gulrajani2017improved} & 18995 & 0.941 & 87.660 & 68.819 & 68.651   \\
%3D-$\alpha$GAN & 46158 & 0.925\\
3D-$\alpha$WGAN \cite{kwon2019generation} & 15713 & 0.984 & 98.853 & 94.814 & 90.102 \\ \hline
Medical Diffusion \cite{khader2022medical} & 16401 & 0.919 & 50.987 & 46.109 & 40.059 \\ \hline
3D-VQGAN-lat4 & 16988 & 0.952 & 44.711 & 34.603 & 32.385    \\
3D-VQGAN-lat8 & \textbf{14982} & \textbf{0.905} & \textbf{26.107} & \textbf{22.686}  & \textbf{15.322}    \\ \hline
Real       & --   & 0.853 & -- & -- & --    
\end{tabular}
\label{quan_res}
\end{table}

Next, we access the model complexity in terms of the total number of trainable parameters and the generation time for one ROI data for all models shown in Table \ref{gen_time}. Notice that the trainable parameters of VQGAN models are the sum of the first-stage autoencoder and the transformer in the second stage. We can see that even though the proposed VQGAN models result in most model parameters, they can still generate one ROI sample in a reasonable time. Compared to the diffusion baseline, our 3D-VQGAN-lat8 model can generate a more realistic ROI sample in less time, which again validates its effectiveness and performance.

\begin{table}[ht]
\centering
\caption{The average image generation time for each model on both GPU and CPU, we also report the trainable parameters for all models.}
\begin{tabular}{c|c|c|c}
                                                            & Params (M) & GPU(s)   & CPU(s)   \\ \hline
3D-WGAN-GP \cite{gulrajani2017improved}                                                  & 49.70 & 1.79 & 6.21 \\
3D-$\alpha$WGAN \cite{kwon2019generation}  & 210.39 & 1.73  & 6.97  \\ \hline
Medical Diffusion \cite{khader2022medical} & 80.97 & 17.63 & 166.85 \\ \hline
3D-VQGAN-lat4                                               & 393.90 & 3.04  & 31.53 \\
3D-VQGAN-lat8                                               & 385.11 & 15.31 & 60.29 \\ \hline
\end{tabular}
\label{gen_time}
\end{table}

\noindent \textbf{Ablation Study. }We conduct an ablation study to validate the effectiveness of using the image gradient loss (IGL) module during the training of the autoencoder in Stage 1. In Section \ref{model_arc}, our hypothesis is that incorporating the IGL will enhance the fidelity of the reconstructed image, ensuring that it retains as many fine details as the original image, and enforcing the codebook features to preserve the important features. Since our 3D-VQGAN-lat8 model yields better performance compared to the 3D-VQGAN-lat4 model shown in Table \ref{quan_res}, we access the reconstruction performance in terms of the image quality by computing the mean Peak Signal-to-Noise Ratio (PSNR) and 3D Structural Similarity Index (3D-SSIM) value \cite{sanchez2018brain} between this model when trained with and without the proposed gradient loss, as shown in Table \ref{img_grad_abl}. By using the proposed IGL, the reconstructed image exhibits enhanced quality and a closer resemblance to the original image based on the higher value of PSNR and 3D-SSIM. These findings suggest the utility of the proposed IGL in preserving fine details within 3D medical images. We have shown superior performance in the image generation task using IGL, and we will further validate it on the classification task in the following sections.

\begin{table}[ht]
\centering
\caption{Ablation result on the image gradient loss (IGL) module with 3D-VQGAN-lat8 model, mean PSNR and 3D-SSIM are computed over the entire LGG dataset}
\begin{tabular}{c|c|c|c}
                      & IGL Module & PSNR $\uparrow$  & 3D-SSIM $\uparrow$ \\ \hline
3D-VQGAN-lat8  & \Checkmark   & \textbf{30.834} & \textbf{0.975}   \\ \hline
3D-VQGAN-lat8  &  \XSolidBrush & 28.223 & 0.960   \\ \hline
\end{tabular}
\label{img_grad_abl}
\end{table}

\subsection{Classification Results on BraTS Dataset} \label{class_res_brats}

Our proposed method addresses the imbalanced training data problem by generating data for the minority class (LGG). To validate its effectiveness, we trained a classification model to distinguish between HGG and LGG brain tumor types. Table \ref{cls_res} presents the classification performance on the standalone test set containing 25 HGG and 25 LGG patients, measured by AUC, F1-score (harmonic mean of Precision and Recall), Accuracy, Precision, and Recall. We use the default probability of 0.5 for all metrics. As can be seen, the result of our proposed method outperforms the baseline results on all metrics. Specifically, our proposed method surpasses the baseline by up to 6.4\% in AUC, 3.4\% in F1-score, and 5.4\% in Accuracy, indicating the proposed method improves the classifier over the model trained with traditional augmentations, and also performs better than the reference model. The classifier that pretrained on synthetic LGGs from our proposed method also outperforms those from the two GAN baselines 3D-WGAN-GP and 3D-$\alpha$WGAN by up to 7.8\%, 2.0\% in AUC; 4.5\%, 8.0\% in F1-score, and 8.5\%, 6.7\% in Accuracy, respectively. Additionally, compared to the state-of-the-art diffusion model, our proposed method still improved by 1.6\% in AUC, 4.6\% in F1-score, and 5.4\% in Accuracy. We also notice that the classifier trained on 3D-VQGAN-lat8 performs better than the one with smaller latent dimensions. This demonstrates images generated by the larger latent feature maps result in better image quality and fidelity which conforms with the quantitative evaluation presented in Section \ref{qua_ana_brats}.

%and Medical Diffusion baseline by 2.0\%, 1.6\% in AUC; 8.0\%, 4.6\% in F1-score, and 6.7\%, 5.4\% in Accuracy, respectively. 

\begin{table}
\centering
\caption{Classification results for all experiments on BraTS dataset. We run all for three trials and report as mean$\pm$standard deviation. Data $(a),(b)$ and $(c)$ refer to Section \ref{exp_res}.}
\begin{adjustwidth}{-1.75cm}{}
\begin{tabular}{c|c|c|c|c|c|c|c|c}
           & Data & Pretrain  & Fine-tune & AUC         & F1-Score & Accuracy & Precision & Recall \\ \hline
Reference  & $(a)$ & \XSolidBrush       & \XSolidBrush        & 0.641$\pm$0.110 & 0.529$\pm$0.223 & 0.566$\pm$0.010 & 0.533$\pm$0.094 & 0.603$\pm$0.202   \\
Traditional Augmentations   & $(b)$  & \XSolidBrush           & \XSolidBrush        & 0.657$\pm$0.028 & 0.634$\pm$0.031 & 0.593$\pm$0.034 & 0.591$\pm$0.066 & 0.720$\pm$0.142 \\ \hline
.3D-WGAN-GP \cite{gulrajani2017improved}  & $(c)$ & \Checkmark    & \Checkmark   & 0.643$\pm$0.081            & 0.623$\pm$0.052 & 0.562$\pm$0.014 & 0.609$\pm$0.052 & 0.653$\pm$0.240     \\
%3D-$\alpha$GAN & Yes & $(c)$ & Yes & 0.648$\pm$0.064 & x \\
3D-$\alpha$WGAN \cite{kwon2019generation} & $(c)$ & \Checkmark & \Checkmark & 0.701$\pm$0.086 & 0.588$\pm$0.093 & 0.580$\pm$0.056 & 0.635$\pm$0.124 & 0.667$\pm$0.277 \\ %\hline
Medical Diffusion \cite{khader2022medical} & $(c)$ & \Checkmark & \Checkmark & 0.705$\pm$0.089 & 0.622$\pm$0.075 & 0.593$\pm$0.024 & 0.588$\pm$0.043 & 0.707$\pm$0.199 \\ \hline
3D-VQGAN-lat4 & $(c)$ & \Checkmark     & \Checkmark       & 0.683$\pm$0.066 & 0.602$\pm$0.065 & 0.600$\pm$0.038 & 0.610$\pm$0.087 & 0.680$\pm$0.246    \\
3D-VQGAN-lat8 & $(c)$ & \Checkmark  & \Checkmark       & \textbf{0.721$\pm$0.026}            & \textbf{0.668$\pm$0.023} & \textbf{0.647$\pm$0.037} & \textbf{0.639$\pm$0.080} & \textbf{0.733$\pm$0.135}
\\ \hline
\end{tabular}
\end{adjustwidth}
\label{cls_res}
\end{table}

\subsection{Results for Generated Images on pLGG Dataset}

\begin{figure}[t]
     \centering
     %\begin{subfigure}[b]{0.32\textwidth}
     \centering
     \includegraphics[width=1.1\textwidth]{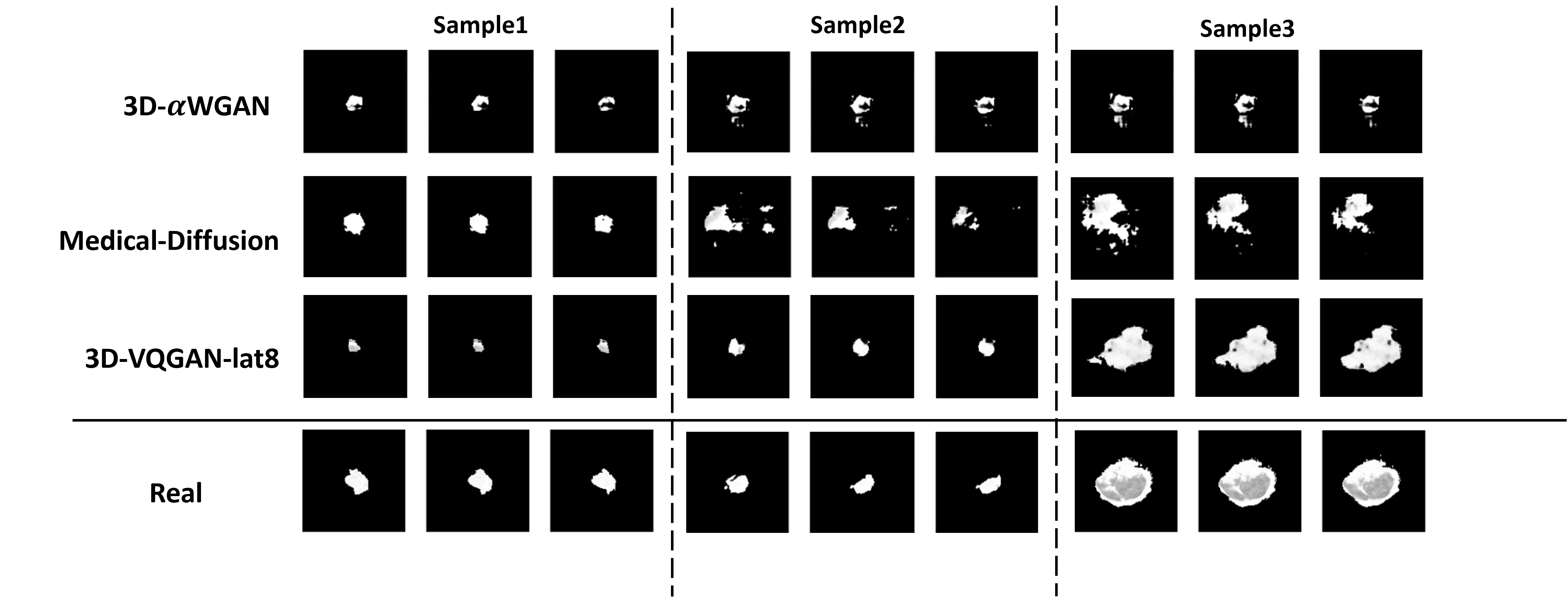}
     \caption{Qualitative comparison between generated and real BRAF V600E Mutation ROIs of three sample tumors. We show the center three slices in the axial plane for each ROI data. The suffix ``lat8'' represents the latent feature map of size $8 \times 8 \times 8$.}
     \label{qual_res_plgg}
\end{figure}

% ... We only show the results for a, b, c models. analysis of qualitative results
Starting from this section, we focus on our internal pLGG dataset and report the qualitative and quantitaive performance on both 3D tumor ROIs generation and classification tasks. We first report the qualitative performance of synthetic BRAF V600E Mutation ROIs generated from both our proposed and baseline models in our internal SickKids pLGG dataset. We have intentionally excluded the 3D-WGAN-GP model \cite{gulrajani2017improved} from all experiments in this section, as it has exhibited a trend to produce blurry images with major artifacts, as demonstrated in Figure \ref{qual_res}. Additionally, we have proven that the 3D-VQGAN-lat8 model is better than the 3D-VQGAN-lat4 model both qualitatively and quantitatively in Section \ref{qua_ana_brats} and \ref{class_res_brats}, so we omit the 3D-VQGAN-lat4 model as well. In Figure \ref{qual_res_plgg}, the generated ROIs for the BRAF V600E Mutation tumor subtype are provided. The 3D-$\alpha$WGAN model produces unrealistic ROIs with unexpected noises, artifacts, and the mode collapse issue is observed. The generated ROIs from the diffusion model are better than 3D-$\alpha$WGAN in terms of the variability of the tumor size, however, they still exhibit noise and unexpected regions. Finally, there are no obvious observations of noisy and unexpected artifacts on the generated ROIs from our proposed method.

\begin{figure}[t]
     \centering
     %\begin{subfigure}[b]{0.32\textwidth}
     \centering
     \includegraphics[width=1.\textwidth]{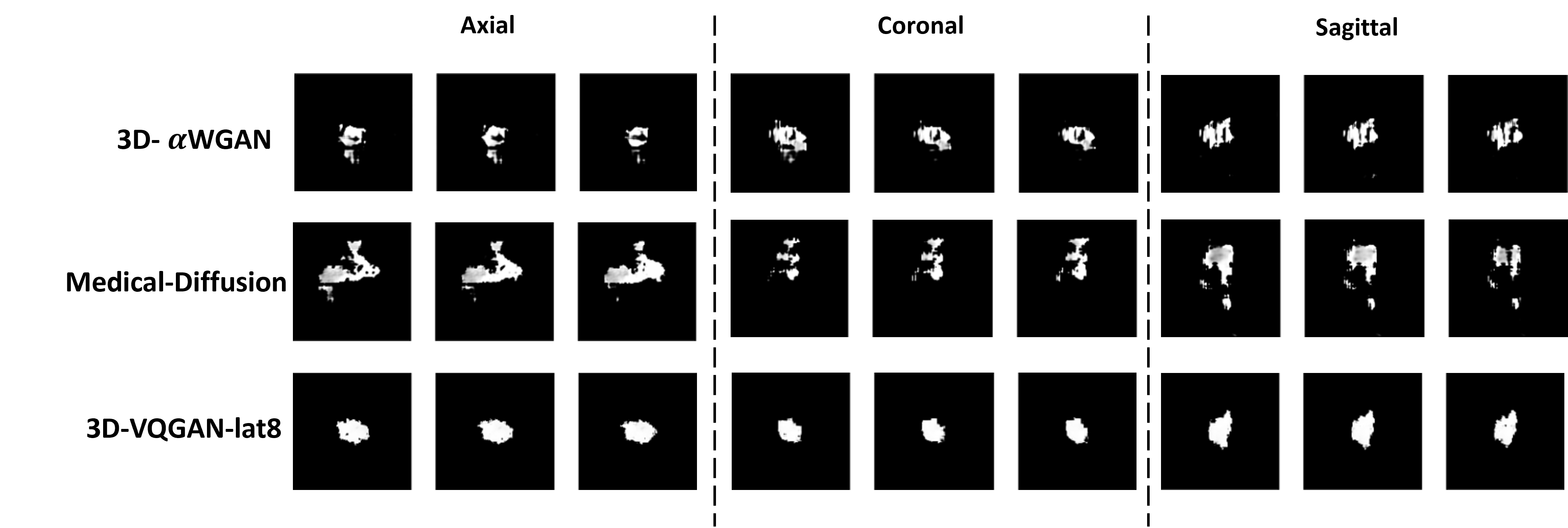}
     \caption{Qualitative comparison between generated BRAF V600E Mutation ROIs from our proposed model and other baseline models visualized in the axial, coronal, and sagittal plane. We show the center three slices. Figures have been enlarged to show the high fidelity of synthetic ROIs generated by our approach.}
     \label{qual_res_acs_plgg}
\end{figure}

\begin{figure}[t]
     \centering
     %\begin{subfigure}[b]{0.32\textwidth}
     \centering
     \includegraphics[width=1\textwidth]{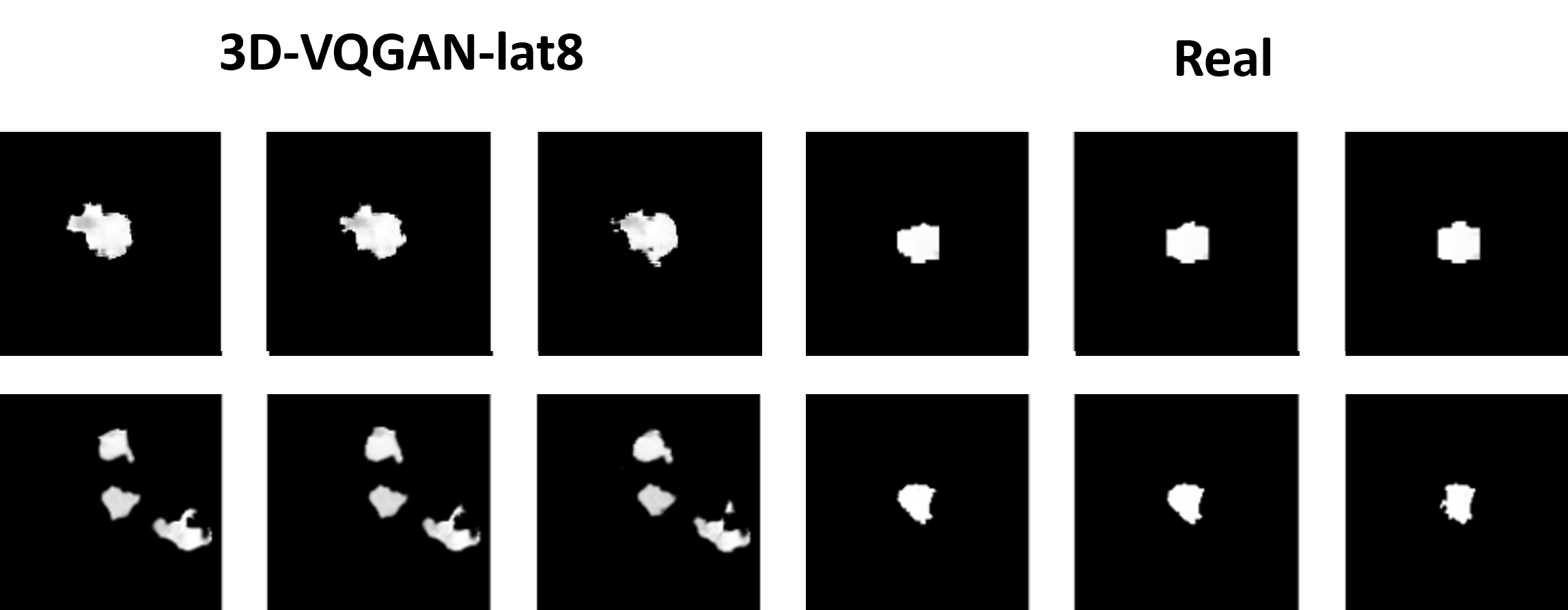}
     \caption{An example of BRAF V600E Mutation ROIs generated by our model and their closest real ROIs based on the 3D-SSIM score. We enlarged the figure for better visualization and only showed the center three slices in the Axial plane.}
     \label{qual_res_mostlike_plgg}
\end{figure}

In Table \ref{quan_res_plgg}, we observe a consistent trend as seen in the previous analysis from Section \ref{qua_ana_brats}: our proposed 3D-VQGAN-lat8 model consistently outperforms all baseline models. The strikethrough values in the 3D-$\alpha$WGAN model indicate the model does not converge properly after excessive exploration of different combinations of hyperparameters. Our proposed method has the closest MS-SSIM score to the one for real data (0.951 vs. 0.942), demonstrating the generated images effectively preserve the pairwise diversity as in the real images. Notice that the MS-SSIM score between real images is relatively high, this may be due to the relatively small tumor sizes in the original dataset. For the MMD score, our model has the lowest value, closely followed by the diffusion baseline. This outcome suggests that our model effectively captures the distribution of real data. Finally, the FID score on three views serves as additional validation of the performance of our model in generating realistic BRAF V600E Mutation ROIs. To provide a visual perspective, we present a sample of generated ROIs in the Axial, Coronal, and Sagittal planes in Figure \ref{qual_res_acs_plgg}, as done in Section \ref{qua_ana_brats}. Additionally, an example of the generated ROIs and their closest real ROIs measured by 3D-SSIM is also shown in Figure \ref{qual_res_mostlike_plgg}. In conclusion, our findings demonstrate that the proposed method can generate realistic BRAF V600E Mutation ROIs, closely aligned with the distribution of real ROIs.

We have chosen not to delve into the model complexity in terms of the image generation time, as we discussed in Section \ref{qua_ana_brats}. The reason for this omission is that the generation times for all models on the pLGG dataset closely mirror those outlined in Table \ref{gen_time} for the BraTS dataset. This similarity can be attributed to the uniformity of the models and their consistent original and latent space image size.

\begin{table}
\centering
\caption{Quantitative results of generated images in the pLGG dataset. Lower values indicate better performance for all scores except MS-SSIM. The MS-SSIM value closest to the real value indicates the best. FID-A, -C, -S refers to calculating the FID over the Axial, Coronal, and Sagittal plane, respectively}
\begin{tabular}{c|c|c|c|c|c}
           & MMD(B=3) $\downarrow$    & MS-SSIM & FID-A $\downarrow$ & FID-C $\downarrow$ & FID-S $\downarrow$  \\ \hline
3D-$\alpha$WGAN \cite{gulrajani2017improved} & \sout{15205} & \sout{0.994} & \sout{237.136} & \sout{267.913} & \sout{222.915}  \\
%3D-$\alpha$GAN & 46158 & 0.925\\
%3D-$\alpha$WGAN \cite{kwon2019generation} & 15713 & 0.984 & 263.996 & 247.080 & 259.178 \\ \hline
Medical Diffusion \cite{khader2022medical} & 14523 & 0.962 & 25.736 & 21.634 & 18.793 \\ \hline
% 3D-VQGAN-lat4 & 16988 & 0.952 & 44.711 & 34.603 & 32.385    \\
3D-VQGAN-lat8 & \textbf{14483} & \textbf{0.951} & \textbf{13.716} & \textbf{10.786}  & \textbf{9.956}    \\ \hline
Real       & --   & 0.942 & -- & -- & --    
\end{tabular}
\label{quan_res_plgg}
\end{table}

\subsection{Classification Results on pLGG Dataset}

Now, we investigate the classification performance of the proposed method for the internal pLGG dataset between BRAF Fusion and BRAF V600E Mutation tumor subtypes. Since the baseline 3D-$\alpha$WGAN model does not converge as illustrated in the previous section and 3D-VQGAN-lat4 is suboptimal to our 3D-VQGAN-lat8 model, we omit the classification results for both models. The metrics are the same as we discussed in Section \ref{class_res_brats}. In Table \ref{cls_res_plgg}, we can see that our 3D-VQGAN-lat8 model still outperforms all other baselines. Specifically, when we use traditional augmentation techniques to form a balanced dataset, the AUC improves by 0.06\%, the F1-score improves by 6.5\%, and the Accuracy improves by 3.3\% compared to the reference model results. However, our proposed method maintains its superiority, outperforming the traditional augmentations by 4.3\% in AUC, 7.3\% in F1-score, and 9.2\% in Accuracy. Finally, compared to our diffusion-based augmentation baseline, our precision score is marginally lower than that of the diffusion model, but all other performance metrics outperform the diffusion model, particularly a 8.2\%, 9.0\%, 5.8\% improvement in AUC, F1-score, and Accuracy, respectively. This finding reinforces our earlier analysis, demonstrating that the proposed method not only generates realistic BRAF V600E Mutation tumor ROIs but also exhibits superior performance compared to other baseline methods.
% This observation aligns with our analysis in the previous section, the proposed method indeed generates realistic BRAF V600E Mutation tumor ROIs and has a superior performance compared to other baseline methods.
% under the imbalanced training data setting (reference model), our model achieves a 4.9\% improvement in AUC, 13.8\% improvement in F1-score, and 12.5\% improvement in Accuracy.

% 0.647$\pm$0.037

\begin{table}
\centering
\caption{Classification results for experiments on the internal pLGG dataset. We run all for three trials and report as mean$\pm$standard deviation. Data $(a),(b)$ and $(c)$ refer to Section \ref{exp_res}.}
\begin{adjustwidth}{-1.75cm}{}
\begin{tabular}{c|c|c|c|c|c|c|c|c}
           & Data  & Pretrain   & Fine-tune & AUC         & F1-Score & Accuracy & Precision & Recall \\ \hline
Reference  & $(a)$ & \XSolidBrush       & \XSolidBrush        & 0.860$\pm$0.025 & 0.720$\pm$0.110 & 0.725$\pm$0.089 & 0.734$\pm$0.108 & 0.767$\pm$0.224    \\
Traditional Augmentations  & $(b)$  & \XSolidBrush          & \XSolidBrush        & 0.866$\pm$0.032 & 0.785$\pm$0.029 & 0.758$\pm$0.042 & 0.764$\pm$0.044 & 0.833$\pm$0.103  \\ \hline
%3D-WGAN-GP \cite{gulrajani2017improved} & \Checkmark      & $(c)$ & \Checkmark   & 0.643$\pm$0.081            & 0.623$\pm$0.052 & 0.562$\pm$0.014      \\
%3D-$\alpha$GAN & Yes & $(c)$ & Yes & 0.648$\pm$0.064 & x \\
%3D-$\alpha$WGAN \cite{kwon2019generation} & \Checkmark & $(c)$ & \Checkmark & 0.701$\pm$0.086 & 0.588$\pm$0.093 & 0.580$\pm$0.056 \\ %\hline
% Medical Diffusion \cite{khader2022medical} & \Checkmark & $(c)$ & \Checkmark & 0.860$\pm$0.013 & 0.834$\pm$0.004 & 0.825$\pm$0.001 \\ \hline
Medical Diffusion \cite{khader2022medical} & $(c)$ & \Checkmark & \Checkmark & 0.827$\pm$0.019 & 0.768$\pm$0.072 & 0.792$\pm$0.031 & \textbf{0.861$\pm$0.099} & 0.733$\pm$0.165 \\ \hline
%3D-VQGAN-lat4 & \Checkmark      & $(c)$ & \Checkmark       & 0.683$\pm$0.066 & 0.602$\pm$0.065 & 0.600$\pm$0.038    \\
3D-VQGAN-lat8 & $(c)$ & \Checkmark  & \Checkmark       & \textbf{0.909$\pm$0.011}            & \textbf{0.858$\pm$0.011} & \textbf{0.850$\pm$0.020} & 0.824$\pm$0.055 & \textbf{0.900$\pm$0.041}
\\ \hline
\end{tabular}
\end{adjustwidth}
\label{cls_res_plgg}
\end{table}

\section{Conclusions}

In this work, we introduce a novel framework based on the VQGAN architecture and the transformer with a masked token modeling strategy to generate realistic 3D tumor ROIs from limited data. We demonstrate the effectiveness and robustness of our method, yielding competitive results in terms of MMD, MS-SSIM, slice-wise FID, and classification performance  across different brain tumor types on two distinct datasets. Our approach opens up possibilities for augmenting rare brain tumor types and facilitating diagnoses using ROIs. Our method also enables a new research avenue of generating only the tumor ROIs instead of the whole brain slices. We have also established a benchmark for tumor ROI generation and classification tasks, laying the foundation for future improvements. With the capability to work with small datasets, our method has the potential to integrate into real clinical routines for brain tumor type classification. However, a limitation of our work lies in its focus on unconditional ROI generation, hence the same model can not be generalized to generate other types of tumors and we have to retrain the model for each new tumor type. Additionally, the quantitative metrics MMD, MS-SSIM, and slice-wise FID could be further improved by combining the multimodal knowledge (i.e., radiology reports) to guide the generation process. In future work, we aim to extend our proposed method into a unified framework capable of conditionally generating ROIs based on class labels and textual information. Lastly, we envision that our proposed method can be adapted for other diseases beyond brain tumors.

% {\color{red} mentions that the proposed method perform well in generating small ROIs, but not good for large ROI}
\section{Acknowledgements:}

This research has been made possible with the financial support of the Canadian Institutes of Health Research (CIHR) (Funding Reference Number: 481135).

\bibliographystyle{splncs04}
\bibliography{ref}

\begin{thebibliography}{10}
\providecommand{\url}[1]{\texttt{#1}}
\providecommand{\urlprefix}{URL }
\providecommand{\doi}[1]{https://doi.org/#1}

\bibitem{ashraf2021deep}
Ashraf, A., Naz, S., Shirazi, S.H., Razzak, I., Parsad, M.: Deep transfer learning for alzheimer neurological disorder detection. Multimedia Tools and Applications pp. 1--26 (2021)

\bibitem{bakas2017advancing}
Bakas, S., Akbari, H., Sotiras, A., Bilello, M., Rozycki, M., Kirby, J.S., Freymann, J.B., Farahani, K., Davatzikos, C.: Advancing the cancer genome atlas glioma mri collections with expert segmentation labels and radiomic features. Scientific data  \textbf{4}(1),  1--13 (2017)

\bibitem{bakas2018identifying}
Bakas, S., Reyes, M., Jakab, A., Bauer, S., Rempfler, M., Crimi, A., Shinohara, R.T., Berger, C., Ha, S.M., Rozycki, M., et~al.: Identifying the best machine learning algorithms for brain tumor segmentation, progression assessment, and overall survival prediction in the brats challenge. arXiv preprint arXiv:1811.02629  (2018)

\bibitem{bao2021beit}
Bao, H., Dong, L., Piao, S., Wei, F.: Beit: Bert pre-training of image transformers. arXiv preprint arXiv:2106.08254  (2021)

\bibitem{de2019management}
de~Blank, P., Bandopadhayay, P., Haas-Kogan, D., Fouladi, M., Fangusaro, J.: Management of pediatric low-grade glioma. Current opinion in pediatrics  \textbf{31}(1), ~21 (2019)

\bibitem{chang2022maskgit}
Chang, H., Zhang, H., Jiang, L., Liu, C., Freeman, W.T.: Maskgit: Masked generative image transformer. In: Proceedings of the IEEE/CVF Conference on Computer Vision and Pattern Recognition. pp. 11315--11325 (2022)

\bibitem{cheng2015enhanced}
Cheng, J., Huang, W., Cao, S., Yang, R., Yang, W., Yun, Z., Wang, Z., Feng, Q.: Enhanced performance of brain tumor classification via tumor region augmentation and partition. PloS one  \textbf{10}(10),  e0140381 (2015)

\bibitem{crowson2022vqgan}
Crowson, K., Biderman, S., Kornis, D., Stander, D., Hallahan, E., Castricato, L., Raff, E.: Vqgan-clip: Open domain image generation and editing with natural language guidance. In: European Conference on Computer Vision. pp. 88--105. Springer (2022)

\bibitem{dar2019image}
Dar, S.U., Yurt, M., Karacan, L., Erdem, A., Erdem, E., Cukur, T.: Image synthesis in multi-contrast mri with conditional generative adversarial networks. IEEE transactions on medical imaging  \textbf{38}(10),  2375--2388 (2019)

\bibitem{devlin2018bert}
Devlin, J., Chang, M.W., Lee, K., Toutanova, K.: Bert: Pre-training of deep bidirectional transformers for language understanding. arXiv preprint arXiv:1810.04805  (2018)

\bibitem{dhariwal2021diffusion}
Dhariwal, P., Nichol, A.: Diffusion models beat gans on image synthesis. Advances in neural information processing systems  \textbf{34},  8780--8794 (2021)

\bibitem{esser2021imagebart}
Esser, P., Rombach, R., Blattmann, A., Ommer, B.: Imagebart: Bidirectional context with multinomial diffusion for autoregressive image synthesis. Advances in neural information processing systems  \textbf{34},  3518--3532 (2021)

\bibitem{esser2021taming}
Esser, P., Rombach, R., Ommer, B.: Taming transformers for high-resolution image synthesis. In: Proceedings of the IEEE/CVF conference on computer vision and pattern recognition. pp. 12873--12883 (2021)

\bibitem{ge2020deep}
Ge, C., Gu, I.Y.H., Jakola, A.S., Yang, J.: Deep semi-supervised learning for brain tumor classification. BMC Medical Imaging  \textbf{20}(1),  1--11 (2020)

\bibitem{ge2022long}
Ge, S., Hayes, T., Yang, H., Yin, X., Pang, G., Jacobs, D., Huang, J.B., Parikh, D.: Long video generation with time-agnostic vqgan and time-sensitive transformer. In: Computer Vision--ECCV 2022: 17th European Conference, Tel Aviv, Israel, October 23--27, 2022, Proceedings, Part XVII. pp. 102--118. Springer (2022)

\bibitem{ghazal2022alzheimer}
Ghazal, T.M., Abbas, S., Munir, S., Khan, M., Ahmad, M., Issa, G.F., Zahra, S.B., Khan, M.A., Hasan, M.K.: Alzheimer disease detection empowered with transfer learning. Computers, Materials \& Continua  \textbf{70}(3) (2022)

\bibitem{gretton2012kernel}
Gretton, A., Borgwardt, K.M., Rasch, M.J., Sch{\"o}lkopf, B., Smola, A.: A kernel two-sample test. The Journal of Machine Learning Research  \textbf{13}(1),  723--773 (2012)

\bibitem{gulrajani2017improved}
Gulrajani, I., Ahmed, F., Arjovsky, M., Dumoulin, V., Courville, A.C.: Improved training of wasserstein gans. Advances in neural information processing systems  \textbf{30} (2017)

\bibitem{han2018gan}
Han, C., Hayashi, H., Rundo, L., Araki, R., Shimoda, W., Muramatsu, S., Furukawa, Y., Mauri, G., Nakayama, H.: Gan-based synthetic brain mr image generation. In: 2018 IEEE 15th international symposium on biomedical imaging (ISBI 2018). pp. 734--738. IEEE (2018)

\bibitem{hao2021transfer}
Hao, R., Namdar, K., Liu, L., Khalvati, F.: A transfer learning--based active learning framework for brain tumor classification. Frontiers in Artificial Intelligence  \textbf{4},  635766 (2021)

\bibitem{hara2017learning}
Hara, K., Kataoka, H., Satoh, Y.: Learning spatio-temporal features with 3d residual networks for action recognition. In: Proceedings of the IEEE international conference on computer vision workshops. pp. 3154--3160 (2017)

\bibitem{harvey2022flexible}
Harvey, W., Naderiparizi, S., Masrani, V., Weilbach, C., Wood, F.: Flexible diffusion modeling of long videos. Advances in Neural Information Processing Systems  \textbf{35},  27953--27965 (2022)

\bibitem{he2022masked}
He, K., Chen, X., Xie, S., Li, Y., Doll{\'a}r, P., Girshick, R.: Masked autoencoders are scalable vision learners. In: Proceedings of the IEEE/CVF conference on computer vision and pattern recognition. pp. 16000--16009 (2022)

\bibitem{heusel2017gans}
Heusel, M., Ramsauer, H., Unterthiner, T., Nessler, B., Hochreiter, S.: Gans trained by a two time-scale update rule converge to a local nash equilibrium. Advances in neural information processing systems  \textbf{30} (2017)

\bibitem{ho2022imagen}
Ho, J., Chan, W., Saharia, C., Whang, J., Gao, R., Gritsenko, A., Kingma, D.P., Poole, B., Norouzi, M., Fleet, D.J., et~al.: Imagen video: High definition video generation with diffusion models. arXiv preprint arXiv:2210.02303  (2022)

\bibitem{ho2020denoising}
Ho, J., Jain, A., Abbeel, P.: Denoising diffusion probabilistic models. Advances in neural information processing systems  \textbf{33},  6840--6851 (2020)

\bibitem{holland2001progenitor}
Holland, E.C.: Progenitor cells and glioma formation. Current opinion in neurology  \textbf{14}(6),  683--688 (2001)

\bibitem{hong20213d}
Hong, S., Marinescu, R., Dalca, A.V., Bonkhoff, A.K., Bretzner, M., Rost, N.S., Golland, P.: 3d-stylegan: A style-based generative adversarial network for generative modeling of three-dimensional medical images. In: Deep Generative Models, and Data Augmentation, Labelling, and Imperfections: First Workshop, DGM4MICCAI 2021, and First Workshop, DALI 2021, Held in Conjunction with MICCAI 2021, Strasbourg, France, October 1, 2021, Proceedings 1. pp. 24--34. Springer (2021)

\bibitem{huang2023not}
Huang, M., Mao, Z., Wang, Q., Zhang, Y.: Not all image regions matter: Masked vector quantization for autoregressive image generation. In: Proceedings of the IEEE/CVF Conference on Computer Vision and Pattern Recognition. pp. 2002--2011 (2023)

\bibitem{johnson2016perceptual}
Johnson, J., Alahi, A., Fei-Fei, L.: Perceptual losses for real-time style transfer and super-resolution. In: Computer Vision--ECCV 2016: 14th European Conference, Amsterdam, The Netherlands, October 11-14, 2016, Proceedings, Part II 14. pp. 694--711. Springer (2016)

\bibitem{khader2022medical}
Khader, F., Mueller-Franzes, G., Arasteh, S.T., Han, T., Haarburger, C., Schulze-Hagen, M., Schad, P., Engelhardt, S., Baessler, B., Foersch, S., et~al.: Medical diffusion--denoising diffusion probabilistic models for 3d medical image generation. arXiv preprint arXiv:2211.03364  (2022)

\bibitem{krishnatry2016clinical}
Krishnatry, R., Zhukova, N., Guerreiro~Stucklin, A.S., Pole, J.D., Mistry, M., Fried, I., Ramaswamy, V., Bartels, U., Huang, A., Laperriere, N., et~al.: Clinical and treatment factors determining long-term outcomes for adult survivors of childhood low-grade glioma: a population-based study. Cancer  \textbf{122}(8),  1261--1269 (2016)

\bibitem{kwon2019generation}
Kwon, G., Han, C., Kim, D.s.: Generation of 3d brain mri using auto-encoding generative adversarial networks. In: Medical Image Computing and Computer Assisted Intervention--MICCAI 2019: 22nd International Conference, Shenzhen, China, October 13--17, 2019, Proceedings, Part III 22. pp. 118--126. Springer (2019)

\bibitem{la2022anatomically}
La~Barbera, G., Boussaid, H., Maso, F., Sarnacki, S., Rouet, L., Gori, P., Bloch, I.: Anatomically constrained ct image translation for heterogeneous blood vessel segmentation. arXiv preprint arXiv:2210.01713  (2022)

\bibitem{li2023mage}
Li, T., Chang, H., Mishra, S., Zhang, H., Katabi, D., Krishnan, D.: Mage: Masked generative encoder to unify representation learning and image synthesis. In: Proceedings of the IEEE/CVF Conference on Computer Vision and Pattern Recognition. pp. 2142--2152 (2023)

\bibitem{lin2017focal}
Lin, T.Y., Goyal, P., Girshick, R., He, K., Doll{\'a}r, P.: Focal loss for dense object detection. In: Proceedings of the IEEE international conference on computer vision. pp. 2980--2988 (2017)

\bibitem{loshchilov2017decoupled}
Loshchilov, I., Hutter, F.: Decoupled weight decay regularization. arXiv preprint arXiv:1711.05101  (2017)

\bibitem{mathieu2015deep}
Mathieu, M., Couprie, C., LeCun, Y.: Deep multi-scale video prediction beyond mean square error. arXiv preprint arXiv:1511.05440  (2015)

\bibitem{menze2014multimodal}
Menze, B.H., Jakab, A., Bauer, S., Kalpathy-Cramer, J., Farahani, K., Kirby, J., Burren, Y., Porz, N., Slotboom, J., Wiest, R., et~al.: The multimodal brain tumor image segmentation benchmark (brats). IEEE transactions on medical imaging  \textbf{34}(10),  1993--2024 (2014)

\bibitem{muller2022diffusion}
M{\"u}ller-Franzes, G., Niehues, J.M., Khader, F., Arasteh, S.T., Haarburger, C., Kuhl, C., Wang, T., Han, T., Nebelung, S., Kather, J.N., et~al.: Diffusion probabilistic models beat gans on medical images. arXiv preprint arXiv:2212.07501  (2022)

\bibitem{mzoughi2020deep}
Mzoughi, H., Njeh, I., Wali, A., Slima, M.B., BenHamida, A., Mhiri, C., Mahfoudhe, K.B.: Deep multi-scale 3d convolutional neural network (cnn) for mri gliomas brain tumor classification. Journal of Digital Imaging  \textbf{33},  903--915 (2020)

\bibitem{namdar2022tumor}
Namdar, K., Wagner, M.W., Kudus, K., Hawkins, C., Tabori, U., Ertl-Wagner, B., Khalvati, F.: Tumor-location-guided cnns for pediatric low-grade glioma molecular biomarker classification using mri. arXiv preprint arXiv:2210.07287  (2022)

\bibitem{pei2020brain}
Pei, L., Vidyaratne, L., Hsu, W.W., Rahman, M.M., Iftekharuddin, K.M.: Brain tumor classification using 3d convolutional neural network. In: Brainlesion: Glioma, Multiple Sclerosis, Stroke and Traumatic Brain Injuries: 5th International Workshop, BrainLes 2019, Held in Conjunction with MICCAI 2019, Shenzhen, China, October 17, 2019, Revised Selected Papers, Part II 5. pp. 335--342. Springer (2020)

\bibitem{peng2022generating}
Peng, W., Adeli, E., Zhao, Q., Pohl, K.M.: Generating realistic 3d brain mris using a conditional diffusion probabilistic model. arXiv preprint arXiv:2212.08034  (2022)

\bibitem{pereira2018automatic}
Pereira, S., Meier, R., Alves, V., Reyes, M., Silva, C.A.: Automatic brain tumor grading from mri data using convolutional neural networks and quality assessment. In: Understanding and Interpreting Machine Learning in Medical Image Computing Applications: First International Workshops, MLCN 2018, DLF 2018, and iMIMIC 2018, Held in Conjunction with MICCAI 2018, Granada, Spain, September 16-20, 2018, Proceedings 1. pp. 106--114. Springer (2018)

\bibitem{pinaya2023generative}
Pinaya, W.H., Graham, M.S., Kerfoot, E., Tudosiu, P.D., Dafflon, J., Fernandez, V., Sanchez, P., Wolleb, J., da~Costa, P.F., Patel, A., et~al.: Generative ai for medical imaging: extending the monai framework. arXiv preprint arXiv:2307.15208  (2023)

\bibitem{pollack2019childhood}
Pollack, I.F., Agnihotri, S., Broniscer, A.: Childhood brain tumors: current management, biological insights, and future directions: Jnspg 75th anniversary invited review article. Journal of Neurosurgery: Pediatrics  \textbf{23}(3),  261--273 (2019)

\bibitem{razavi2019generating}
Razavi, A., Van~den Oord, A., Vinyals, O.: Generating diverse high-fidelity images with vq-vae-2. Advances in neural information processing systems  \textbf{32} (2019)

\bibitem{rehman2020deep}
Rehman, A., Naz, S., Razzak, M.I., Akram, F., Imran, M.: A deep learning-based framework for automatic brain tumors classification using transfer learning. Circuits, Systems, and Signal Processing  \textbf{39},  757--775 (2020)

\bibitem{rosca2017variational}
Rosca, M., Lakshminarayanan, B., Warde-Farley, D., Mohamed, S.: Variational approaches for auto-encoding generative adversarial networks. arXiv preprint arXiv:1706.04987  (2017)

\bibitem{ryall2020integrated}
Ryall, S., Zapotocky, M., Fukuoka, K., Nobre, L., Stucklin, A.G., Bennett, J., Siddaway, R., Li, C., Pajovic, S., Arnoldo, A., et~al.: Integrated molecular and clinical analysis of 1,000 pediatric low-grade gliomas. Cancer cell  \textbf{37}(4),  569--583 (2020)

\bibitem{sajjad2019multi}
Sajjad, M., Khan, S., Muhammad, K., Wu, W., Ullah, A., Baik, S.W.: Multi-grade brain tumor classification using deep cnn with extensive data augmentation. Journal of computational science  \textbf{30},  174--182 (2019)

\bibitem{sanchez2018brain}
S{\'a}nchez, I., Vilaplana, V.: Brain mri super-resolution using 3d generative adversarial networks. arXiv preprint arXiv:1812.11440  (2018)

\bibitem{shin2018medical}
Shin, H.C., Tenenholtz, N.A., Rogers, J.K., Schwarz, C.G., Senjem, M.L., Gunter, J.L., Andriole, K.P., Michalski, M.: Medical image synthesis for data augmentation and anonymization using generative adversarial networks. In: Simulation and Synthesis in Medical Imaging: Third International Workshop, SASHIMI 2018, Held in Conjunction with MICCAI 2018, Granada, Spain, September 16, 2018, Proceedings 3. pp. 1--11. Springer (2018)

\bibitem{srinivas2022deep}
Srinivas, C., KS, N.P., Zakariah, M., Alothaibi, Y.A., Shaukat, K., Partibane, B., Awal, H., et~al.: Deep transfer learning approaches in performance analysis of brain tumor classification using mri images. Journal of Healthcare Engineering  \textbf{2022} (2022)

\bibitem{sturm2017pediatric}
Sturm, D., Pfister, S.M., Jones, D.T.: Pediatric gliomas: current concepts on diagnosis, biology, and clinical management. Journal of Clinical Oncology  \textbf{35}(21),  2370--2377 (2017)

\bibitem{subramaniam2022generating}
Subramaniam, P., Kossen, T., Ritter, K., Hennemuth, A., Hildebrand, K., Hilbert, A., Sobesky, J., Livne, M., Galinovic, I., Khalil, A.A., et~al.: Generating 3d tof-mra volumes and segmentation labels using generative adversarial networks. Medical Image Analysis  \textbf{78},  102396 (2022)

\bibitem{swati2019brain}
Swati, Z.N.K., Zhao, Q., Kabir, M., Ali, F., Ali, Z., Ahmed, S., Lu, J.: Brain tumor classification for mr images using transfer learning and fine-tuning. Computerized Medical Imaging and Graphics  \textbf{75},  34--46 (2019)

\bibitem{tak2023noninvasive}
Tak, D., Ye, Z., Zapaischykova, A., Zha, Y., Boyd, A., Vajapeyam, S., Chopra, R., Hayat, H., Prabhu, S., Liu, K.X., et~al.: Noninvasive molecular subtyping of pediatric low-grade glioma with self-supervised transfer learning. medRxiv pp. 2023--08 (2023)

\bibitem{tandel2020multiclass}
Tandel, G.S., Balestrieri, A., Jujaray, T., Khanna, N.N., Saba, L., Suri, J.S.: Multiclass magnetic resonance imaging brain tumor classification using artificial intelligence paradigm. Computers in Biology and Medicine  \textbf{122},  103804 (2020)

\bibitem{tudosiu2022morphology}
Tudosiu, P.D., Pinaya, W.H.L., Graham, M.S., Borges, P., Fernandez, V., Yang, D., Appleyard, J., Novati, G., Mehra, D., Vella, M., et~al.: Morphology-preserving autoregressive 3d generative modelling of the brain. In: International Workshop on Simulation and Synthesis in Medical Imaging. pp. 66--78. Springer (2022)

\bibitem{ullah2022effective}
Ullah, N., Khan, J.A., Khan, M.S., Khan, W., Hassan, I., Obayya, M., Negm, N., Salama, A.S.: An effective approach to detect and identify brain tumors using transfer learning. Applied Sciences  \textbf{12}(11), ~5645 (2022)

\bibitem{van2017neural}
Van Den~Oord, A., Vinyals, O., et~al.: Neural discrete representation learning. Advances in neural information processing systems  \textbf{30} (2017)

\bibitem{villa20182016}
Villa, C., Miquel, C., Mosses, D., Bernier, M., Di~Stefano, A.L.: The 2016 world health organization classification of tumours of the central nervous system. La Presse M{\'e}dicale  \textbf{47}(11-12),  e187--e200 (2018)

\bibitem{volokitin2020modelling}
Volokitin, A., Erdil, E., Karani, N., Tezcan, K.C., Chen, X., Van~Gool, L., Konukoglu, E.: Modelling the distribution of 3d brain mri using a 2d slice vae. In: Medical Image Computing and Computer Assisted Intervention--MICCAI 2020: 23rd International Conference, Lima, Peru, October 4--8, 2020, Proceedings, Part VII 23. pp. 657--666. Springer (2020)

\bibitem{wang2013understanding}
Wang, Y., Jiang, T.: Understanding high grade glioma: molecular mechanism, therapy and comprehensive management. Cancer letters  \textbf{331}(2),  139--146 (2013)

\bibitem{wolleb2022diffusion}
Wolleb, J., Bieder, F., Sandk{\"u}hler, R., Cattin, P.C.: Diffusion models for medical anomaly detection. In: International Conference on Medical image computing and computer-assisted intervention. pp. 35--45. Springer (2022)

\bibitem{yang2022diffusion}
Yang, R., Srivastava, P., Mandt, S.: Diffusion probabilistic modeling for video generation. arXiv preprint arXiv:2203.09481  (2022)

\bibitem{youssef2020lower}
Youssef, G., Miller, J.J.: Lower grade gliomas. Current neurology and neuroscience reports  \textbf{20}, ~1--9 (2020)

\bibitem{yu20183d}
Yu, B., Zhou, L., Wang, L., Fripp, J., Bourgeat, P.: 3d cgan based cross-modality mr image synthesis for brain tumor segmentation. In: 2018 IEEE 15th international symposium on biomedical imaging (ISBI 2018). pp. 626--630. IEEE (2018)

\bibitem{yu2021vector}
Yu, J., Li, X., Koh, J.Y., Zhang, H., Pang, R., Qin, J., Ku, A., Xu, Y., Baldridge, J., Wu, Y.: Vector-quantized image modeling with improved vqgan. arXiv preprint arXiv:2110.04627  (2021)

\end{thebibliography}
\end{document}